\documentclass[11pt]{IEEEtran}
\usepackage{times}
\usepackage{url}
\usepackage{authblk}

\usepackage{cite}
\usepackage{hyperref}

  \usepackage[pdftex]{graphicx}

\usepackage{amsmath}
\usepackage{cleveref}

\usepackage{algorithmic}

\usepackage[caption=false,font=normalsize,labelfont=sf,textfont=sf]{subfig}

\usepackage{multirow}
\usepackage{tabularx}
\usepackage{booktabs}
\usepackage{bm}
\usepackage{amssymb}
\usepackage{amsthm}
\usepackage{array}
\theoremstyle{definition}

\usepackage[ruled]{algorithm2e} 
\SetAlCapNameFnt{\small}
\SetAlCapFnt{\small}
\usepackage{amssymb,amsmath}
\usepackage[hang,flushmargin]{footmisc}
\usepackage{lipsum}
\usepackage{makecell}
\usepackage{threeparttable}
\usepackage{color,xcolor}
\makeatletter
\newcommand{\algorithmfootnote}[2][\footnotesize]{%
  \let\old@algocf@finish\@algocf@finish
  \def\@algocf@finish{\old@algocf@finish
    \leavevmode\rlap{\begin{minipage}{\linewidth}
    #1#2
    \end{minipage}}%
  }%
}
\makeatother

\definecolor{version_1}{rgb}{0,0,0}      
\definecolor{version_2}{rgb}{0,0,0}     
\definecolor{version_3}{rgb}{0,0,0} 
\definecolor{version_4}{rgb}{0,0,1}

\begin{document}

\title{Nearest-Better Network for Visualizing and Analyzing Combinatorial Optimization Problems: A Unified Tool}


\author{Yiya~Diao, Changhe~Li$^*$,
        Sanyou~Zeng,
        Xinye~Cai,
        Wenjian~Luo,
        Shengxiang~Yang,
        and Carlos~A.~Coello~Coello,~\IEEEmembership{Fellow,~IEEE}%
\thanks{This work has been accepted for publication in \textit{IEEE Transactions on Evolutionary Computation}. The final published version will be available via IEEE Xplore.}%
\thanks{Y. Diao and W. Luo are with Guangdong Provincial Key Laboratory of Novel Security Intelligence Technologies, School of Computer Science and Technology, Harbin Institute of Technology, Shenzhen 518055, China (e‑mail: diaoyiyacug@gmail.com; luowenjian@hit.edu.cn).}%
\thanks{C. Li and X. Cai are with the School of Artificial Intelligence, Anhui University of Science and Technology, Hefei 231131, China, and also with the State Key Laboratory of Digital Intelligent Technology for Unmanned Coal Mining, Huainan 232001, China (e‑mail: changhe.lw@gmail.com).}%
\thanks{S. Zeng is with the School of Mechanical Engineering and Electronic Information, China University of Geosciences, Wuhan, China (e‑mail: sanyouzeng@gmail.com).}%
\thanks{S. Yang is with the School of Computer Science and Informatics, De Montfort University, Leicester LE1 9BH, U.K.}%
\thanks{C. A. Coello Coello is with the Department of Computer Science, CINVESTAV‑IPN, Mexico City 07360, Mexico, and also (on sabbatical leave) with the School of Engineering and Sciences, Tecnológico de Monterrey, Monterrey, N.L., Mexico.}}%

\maketitle

\begin{abstract}
The Nearest-Better Network (NBN) is a powerful method to visualize sampled data for continuous optimization problems while preserving multiple landscape features. 
However, the calculation of NBN is very time-consuming, and the extension of the method to combinatorial optimization problems is challenging but very important for analyzing the algorithm's behavior. This paper provides a straightforward theoretical derivation showing that the NBN network essentially functions as the maximum probability transition network for algorithms.  This paper also presents an efficient NBN computation method with logarithmic linear time complexity to address the time-consuming issue. 
By applying this efficient NBN algorithm to the OneMax problem and the Traveling Salesman Problem (TSP), we have made several remarkable discoveries for the first time: The fitness landscape of OneMax exhibits neutrality, ruggedness, and modality features. The primary challenges of TSP problems are ruggedness, modality, and deception. Two state-of-the-art TSP algorithms (i.e., EAX and LKH) have limitations when addressing challenges related to modality and deception, respectively. LKH, based on local search operators, fails when there are deceptive solutions near global optima.  
EAX, which is based on a single population, can efficiently maintain diversity. 
However, when multiple attraction basins exist, EAX retains individuals within multiple basins simultaneously, reducing inter-basin interaction efficiency and leading to algorithm's stagnation.

\end{abstract}

\author{
Yiya Diao, Changhe Li$^*$, Sanyou Zeng, Xinye Cai, Wenjian Luo, Shengxiang Yang, and Carlos A. Coello Coello%
\thanks{
This work was supported in part by the National Natural Science Foundation of China under Grant 62476006 and U23B2058, 
in part by the Hubei Provincial Natural Science Foundation of China under Grant 2023AFA049, 
and in part by the Fundamental Research Funds of the AUST under Grant 2024JBZD0007. 
(\textit{Corresponding author: Changhe Li.})
}%
\thanks{
Y. Diao and W. Luo are with the Guangdong Provincial Key Laboratory of Novel Security Intelligence Technologies, School of Computer Science and Technology, Harbin Institute of Technology, Shenzhen, Guangdong 518055, China (e-mail: diaoyiyacug@gmail.com; luowenjian@hit.edu.cn).
}%
\thanks{
C. Li and X. Cai are with the School of Artificial Intelligence, Anhui University of Science and Technology, Hefei 231131, China, and also with the State Key Laboratory of Digital Intelligent Technology for Unmanned Coal Mining, Anhui University of Science and Technology, Huainan 232001, China (e-mail: changhe.lw@gmail.com).
}%
\thanks{
S. Zeng is with the School of Mechanical Engineering and Electronic Information, China University of Geosciences, Wuhan, China (e-mail: sanyouzeng@gmail.com).
}%
\thanks{
S. Yang is with the School of Computer Science and Informatics, De Montfort University, Leicester LE1 9BH, U.K.
}%
\thanks{
C. A. Coello Coello is with the Department of Computer Science, CINVESTAV-IPN, Mexico City 07360, Mexico, and also (on sabbatical leave) with the School of Engineering and Sciences, Tecnologico de Monterrey, Monterrey, N.L., Mexico.
}
}

\section*{Notice}
This work has been accepted for publication in \textit{IEEE Transactions on Evolutionary Computation}. The final published version will be available via IEEE Xplore.

\section{Introduction}
\label{sec:introducton}

Combinatorial optimization problems are a crucial category of optimization problems, prevalent in various real-world applications.
The Traveling Salesman Problem (TSP) is a classic example of a combinatorial optimization problem, and many practical combinatorial optimization problems can be addressed using the TSP model, 
such as vehicle routing problems\:\cite{Laporte1992vehicle}, DNA sequencing\:\cite{Pop2009genome}, and computer chip layout design\:\cite{Alpert2008handbook}.
The study of TSP is of significant importance to both academia and industry.

 There are numerous optimization algorithms for solving the TSP, from which Lin-Kernighan-Helsgaun (LKH)\:\cite{Tinos2018efficient} and Edge Assembly Crossover based GA (EAX)\:\cite{Nagata2013powerful} are two state-of-the-art algorithms. 
 However, research on TSP algorithms appears to have reached a bottleneck in recent years, with few new algorithms showing significant performance improvements\:\cite{Scholz2019genetic}. 
 But, is this really the case? As shown in experiments  reported in\:\cite{Liu2023good}, 
 even for relatively simple TSP instances with 500 cities, 
 both EAX and LKH fail to achieve 100\% accuracy. This indicates that certain landscape features pose challenges that these algorithms struggle to overcome. 
 This also suggests that there is still room for improvement.
 What we truly lack is a robust tool for analyzing combinatorial problems and algorithms. 
 Such tools would enable researchers to effectively identify the inherent difficulties in the problems, pinpoint the algorithms' weaknesses, and would allow to systematically improve existing algorithms.

Fitness Landscape Analysis (FLA) methods aim to analyze and visualize the fitness landscape through various sampling methods. 
These methods assist in landscape feature analysis, algorithm performance analysis, and algorithm design. 
In theory, a good visualization method can help us observe the global and local structure of a fitness landscape as well as the search behavior of algorithms, 
which helps to design efficient algorithms for different types of problems. 
However, visual FLA methods for combinatorial optimization problems are scarce, 
with only three methods currently available:  Low-Dimensional Euclidean Embedding (LDEE)\:\cite{Michalak2019Low}, Local
Optima Network (LON)\:\cite{Ochoa2018Mapping}, and Nearest-Better Network (NBN)\:\cite{Diao2023Nearest}. 
Previous experiments\:\cite{Diao2024Nearest} have shown that NBN can display many landscape features in visualization. 
However, NBN lacks an efficient calculation method to handle the large number of samples generated by combinatorial algorithms and also lacks a theoretical analysis to substantiate its mechanisms.
Based on this, this paper further studies NBN for combinatorial problems. The main contributions of this paper are the following:
\begin{itemize}
  \item This paper attempts to explain the working mechanism of NBN from the perspective of algorithm search behaviors with theoretical analysis. Our preliminary analysis shows that  NBN fundamentally represents the maximum transition probability model of an algorithm.
  This analysis reveals that NBN statistically models the algorithm's behavior, thereby simplifying the original fitness landscape. Moreover, it explains why NBN can preserve most features of the fitness landscape, which have significant impacts on algorithm's performance.
  \item This paper proposes an efficient algorithm to compute NBN for combinatorial optimization problems so that we can visualize the NBN of combinatorial optimization problems in logarithmic linear time complexity.

 
  \item This paper illustrates the fitness landscape structures of combinatorial optimization problems with different landscape features using a tunable black-box discrete optimization benchmarking problem, the W-Model \:\cite{Weise2018difficult}. 
  By adjusting parameters, the NBN network of different W-Model  instances reveals that the W-Model  problem exhibits ruggedness, neutrality, and multimodal features.
  \item This paper conducts an in-depth analysis of the current leading TSP algorithms, EAX and LKH.
Our experimental results reveal that both EAX and LKH have limitations when addressing challenges related to modality and deception, respectively.
Specifically, LKH, which relies on local search operators, struggles with deceptive solutions near the global optima. On the other hand, EAX, being a single-population-based algorithm, excels at maintaining diversity. However, it faces stagnation issues when dealing with multiple basins of attraction due to reduced interaction efficiency between individuals in different basins. 
\end{itemize}

The remainder of this paper is organized as follows. 
Section\:\ref{sec:relatedWork} gives an overview of previous related work.
Section\:\ref{sec:NBN} provides a theoretical proof of NBN.
Section\:\ref{sec:NBN_calculation} details the algorithm to calculate NBN for the given problems.
Section\:\ref{sec:NBN_analysis} presents the experimental analysis of the landscape features for OneMax and TSP, as well as the behavior of algorithms applied to TSP.
Finally, our conclusions  and some potential paths for future research are given in Section\:\ref{sec:conclusion}.

\section{Previous related work}\label{sec:relatedWork}

A general view of fitness landscapes was proposed in\:\cite{Stadler2002Fitness}, in which the fitness landscape consists of three elements $(\boldsymbol X, \chi , f)$:
\begin{itemize}
  \item A set $\boldsymbol{X}$ of potential solutions to the problem,
  \item a notion $\chi$ of neighborhood, nearness, distance, or accessibility on $\boldsymbol X$, and
  \item a fitness function $f: \boldsymbol X  \to \boldsymbol R$. 
  The fitness of a solution indicates how good the solution is, and the larger the fitness value, the better the solution. 
\end{itemize}

There are several definitions related to the fitness landscape and the algorithm's behavior, including the following:
\begin{itemize}
  \item {\bf Search space:} The search space $\boldsymbol{X}$ is the union of all possible solutions of an optimization problem.
  \item {\bf Neighborhood:} The neighborhood relationship is a mapping $\chi $: $ \boldsymbol X \to \boldsymbol N$,
which associates each solution $\mathbf {x}$ with a set of candidate solutions $\boldsymbol N( \mathbf{x})$,

     \item {\bf Basin of Attraction} (BoA) and {\bf local optimum}: 
  $\boldsymbol B(\mathbf x^*)  = \left\{ \mathbf x \in \boldsymbol X   \mid \mathbf x^*=     \text{local-search}(\mathbf x)  \right\}  $, where   the BoA $\boldsymbol B(\mathbf x^*)$ of a local optimum
 $\mathbf{x}^*$ is the set of solutions $\boldsymbol B(\mathbf x^*)$  that approaches $\mathbf x^*$ by utilizing a local search strategy among the decision variable space $\boldsymbol X$\:\cite{ZouCL2022Survey}.

 \item {\bf Search trajectory $\boldsymbol T$}:
  Search trajectory is a sequence of the solutions generated by the algorithm in one run. In this paper,  $\boldsymbol T$ represents the set of solutions of a search trajectory in one run of the algorithm.

\end{itemize}

In the past few decades, numerous FLA methods have been developed, 
from understanding the fitness landscape to guiding the search process. 
This section focuses on the FLA methods for combinatorial optimization problems. These methods can be categorized into non-visual and visual methods based on their ability to visualize data.

\subsection{Non-visual methods}

Non-visualization methods are used to analyze landscape features or to design benchmarks with specific features, 
comparing algorithms' performance on these problems to indirectly evaluate the algorithms' capability to address these features. 
The main non-visualization methods include metric analysis and benchmark design. 
These methods provide researchers with an indirect way to analyze algorithm's behavior, particularly when direct observation of the algorithm's behavior is not available, 
helping to gain deeper insights into algorithm's performance and challenges in real-world problem-solving scenarios.

\subsubsection{Metric analysis}

These methods propose a series of metrics to describe specific features of problems or algorithms. 
Through these metrics, the performance of algorithms under different features can be evaluated, 
and the effectiveness of algorithms in solving problems with similar features can be inferred. Lip used the correlation length to evaluate ruggedness:\cite{Lipsitch1991Adaptation}. 
Davidor employed epistasis variance to assess epistasis\:\cite{Davidor1991Epistasis}. 
Reidys and Stadler utilized a neutral walk to evaluate neutrality\:\cite{Reidys2001Neutrality}. 
Lunacek proposed the dispersion metric to evaluate global topology or the presence of funnels\:\cite{LunacekW2006Dispersion}.
In theory, these metrics can be directly applied to combinatorial optimization problems. However, due to the lack of observable or quantifiable combinatorial benchmarks, their performance has not been validated in this domain.


\subsubsection{Benchmarks design}
These methods involve designing a set of benchmarks with specific landscape features to evaluate algorithm's performance. 
Benchmarks may include instances with different structures or landscape features. 
By comparing how algorithms solve these problems, we can reveal their adaptability and limitations across different features.


In the field of combinatorial optimization, there are very few of such benchmarks available. 
For continuous problems, we can typically verify if benchmarks exhibit the intended features through observation of the two-dimensional continuous problems~\cite{Li2019open}. However, evaluating whether the designed benchmarks accurately reflect the designed features for combinatorial problems is a challenging problem. 
W-Model ~\cite{Weise2018difficult} is the only combinatorial benchmark that allows adjustments of ruggedness, neutrality, and epistasis features.


The effectiveness of the design of the benchmarks depends on whether the set of designed benchmarks appropriately covers the range and variations of target features.
A poorly designed or incomplete benchmark set may lead to misjudgments of the algorithm's behavior or biased analyses.


\subsection{Visual methods}
Theoretically, a good visualization method can help us observe both the global structure and the local structure of the fitness landscape, 
as well as the search behavior of an algorithm. This helps us to improve our understanding of the problem structure and the algorithm's working mechanism as well as to design efficient algorithms.

Visual FLAs for combinatorial problems are very few: LON~\cite{Ochoa2014Local}, LDEE~\cite{Michalak2019Low}, and NBN~\cite{Diao2023Nearest}. The neighborhood relationships between solutions in the fitness landscape of combinatorial optimization problems are extremely complex. For instance, in a TSP problem with 500 cities, if the neighborhood is defined based on 2-opt, a single solution would have approximately $C^2_{500}-1 \approx 124,251 $  neighboring solutions. One critical challenge in visualizing combinatorial optimization problems is how to simplify these neighborhood relationships between solutions.

LON visualize the connections between local optima in the fitness landscape. 
This novel method displays the fitness landscape in the form of a graph where nodes represent local optima and edges indicate the transitions between optima given a specific search operator. 
It uses a 2-opt local operator to search for local optima and a 4-opt operator to escape from the current local optima, 
constructing the local optima network in TSP. Since algorithms consist of different search operators, this method can associate the algorithm's behavior with the structure of the fitness landscape.
LDEE  shows the dynamics of the population for combinatorial problems by mapping combinatorial solutions into a two-dimensional space using a t-distributed stochastic neighbor embedding method. 

NBN can visualize data from any sampling source. 
Previous experiments~\cite{Diao2024Nearest} have shown that NBN can display many features of the landscape in its visualization.
In this paper, we attempt to use NBN to visualize the global structure, as well as the local structure of the fitness landscape, and the search behavior of algorithms for two typical combinatorial optimization problems, i.e., OneMax and TSP. Through NBN's visualization, we try to uncover some unknown difficulties of OneMax and TSP.

\section{Nearest-better Network}\label{sec:NBN}

In this section, we provide a straightforward theoretical analysis to explain why NBN is effective.

To analyze the original fitness landscape, 
the first challenge is to find a method that can handle problems with a huge number of solutions. It is an intuitive idea to partition the original search space into several subspaces. A similar method, named cell mappings techniques\:\cite{Hsu2013Cell}, is used to analyze the global behavior of nonlinear dynamical systems.


Let's assume that ${\boldsymbol X}_N = \left\{ \mathbf x_1,...,\mathbf x_N \right\}$ 
is a big set of sampled solutions from the search space, 
where  $N$ is a large number,
and $\boldsymbol X_N$ approximates the whole search space in this paper, $\boldsymbol X  \rightarrow     \boldsymbol X_N$.
Now, the fitness landscape is simplified to a set consisting of a finite number of solutions, 
but the neighborhood relationships between every two solutions are still complex and unknown, and they still need to be simplified. 



\subsection{A simple format of evolutionary algorithms}
The FLAs aim to help to design efficient algorithms, 
so it is a natural way to analyze the fitness landscape from the perspective of optimization algorithms 
and the previous section also shows that this idea helps to simplify the neighborhood relationship.

There is a considerable number of optimization algorithms, 
and it is practically impossible to analyze them all. 
Here, we consider a simple format of an evolutionary algorithm, a (1 + 1)-ES version, 
with a Gaussian mutation operator.
\begin{equation}
  \label{equ:Alg_generation}
 x'_i \leftarrow x_i + \mathcal{N}(0,r),
  \end{equation}
  where $\mathbf x = \left[ x_1, x_2,...,x_D \right]$, $D$ is the dimensionality of the problem, 
  and $r$ is the mutation step-size.
  \begin{equation}
    \label{equ:Alg_selection}
    \mathbf x  = \left\{
    \begin{aligned}
    &\mathbf x'   & \text{if}\: f(\mathbf x' ) >  f(\mathbf x)  \\
    &\mathbf x    & \text{otherwise}
    \end{aligned}
    \right.
    \end{equation}
Although this type of EA is very simple, most popular EAs share similarities with it.
For example, the Covariance Matrix adaptation Evolution Strategy, (CMA-ES)\:\cite{Hansen2001completely} for continuous optimization problems has a similar format combined with its gradient calculation method. Additionally, the inner mechanism of the powerful LKH for the TSP is also similar to this type of EA.

\subsection{Maximum Transition Network}
\begin{figure}[t]
  \centering
  \includegraphics[width=0.35\textwidth]{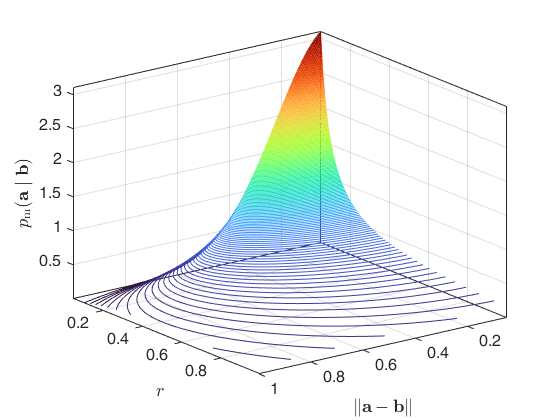}
  \caption{Mutation probability function}
  \label{fig:mutation_pro}
\end{figure}

Then, we can calculate the transition probability between two solutions based on 
Eqs.\:\eqref{equ:Alg_generation} and \eqref{equ:Alg_selection}. Let's assume that the mutations of each dimension are independent and identically distributed. Then, the mutation probability between two solutions
$\mathbf a$ and $\mathbf b$,  
$p_{\text{m}}(\mathbf a \leftarrow \mathbf b)$, is calculated as follows:
\begin{equation}
\begin{aligned}
 & p_{\text{m}}(\mathbf a \leftarrow \mathbf b) \\
  = &p(a_{1}-b_{1}) p(a_{2}-b_{2}) ... p(a_{D}-b_{D}) \\
  = &(2\pi r)^{-\frac{D}{2}} \exp   ( - \frac{\| \mathbf a , \mathbf b \|}{2r}),
\end{aligned}
\end{equation}
Fig.\:\ref{fig:mutation_pro} shows the mutation probability function $p_{\text{m}}(\mathbf a \leftarrow \mathbf b)$
associated with $\| \mathbf a , \mathbf b \|$ and $r$. 
Generally speaking, the mutation step-size $r$ is a pre-defined parameter. 

The selection probability $p_{\text{s}}(\mathbf a \leftarrow \mathbf b)$ is calculated by
\begin{equation}
  \label{equ:Alg_selection_pro}
  p_{\text{s}}(\mathbf a \leftarrow \mathbf b) = \left\{
  \begin{aligned}
  &1,  & \text{if}\: f(\mathbf a ) > f(\mathbf b )  \\
  &0 ,   & \text{otherwise}
  \end{aligned}
  \right.,
  \end{equation}
Finally, the transition probability between the two solutions
 $p(\mathbf a \leftarrow \mathbf b)$ is calculated by
 \begin{equation}
\begin{aligned}
  &p(\mathbf a \leftarrow \mathbf b )\\
   =&  p_{\text{m}}(\mathbf a \leftarrow \mathbf b)  
   p_{\text{s}}(\mathbf a \leftarrow \mathbf b) \\
    =& 
    \left\{
   \begin{aligned}
   &(2\pi r)^{-\frac{D}{2}}  \exp   ( - \frac{\| \mathbf a ,\mathbf b \|^2}{2r}),  
   & \text{if}\: f (\mathbf a ) > f(\mathbf b )  \\
   &0,    & \text{otherwise}
   \end{aligned} 
   \right.
  \end{aligned}
\end{equation}

Now that the relationship between every two solutions is known to us, theoretically, we can analyze the fitness landscape according to these equations. 
However, if we consider all the relationships, 
the fitness landscape analysis will be too complex to be applied to any high-dimensional or combinatorial problem.
So, we try to simplify the relationship by maintaining only the maximum transition relationship for each solution.
In the network, $ \mathbf {\beta} ( \mathbf x) $  is the solution with maximum transition from solution $\mathbf x$, 
which is defined as:
\begin{equation}
\begin{aligned}
  \mathbf{\beta} (\mathbf x)
   &=   \arg \max_{\mathbf y \in \boldsymbol X_N} p(\mathbf y \leftarrow \mathbf x) \\
  &=    \arg \min_{\mathbf y \in  \left\{   \mathbf y  \mid  \mathbf y \in \boldsymbol X_N,  f(\mathbf y ) > f(\mathbf x )  \right\} } \| \mathbf y, \mathbf x \| 
\end{aligned}
\end{equation}
$\mathbf {\beta} ( \mathbf x) $  is also known as the nearest better solution in\:\cite{Preuss2010niching}.
Note that for the global optimum $\mathbf o$, there is no better solution.

With the simplification of the original fitness landscape and  the nearest better relationship, 
the maximum transition network can be defined as a directed graph $\mathbf G  =  (\boldsymbol V, \boldsymbol E)  $, 
where the set of vertices is the set of representative solutions,   $\boldsymbol V = \boldsymbol X_N$, 
and the set of edges is the nearest better relationship for every solution,
$\boldsymbol E= \left\{ (\mathbf x, \mathbf{\beta } (\mathbf x)) \mid \mathbf x \in \boldsymbol X_N \right\} $.

\begin{figure}[t]
  \centering
  \includegraphics[width=0.35\textwidth]{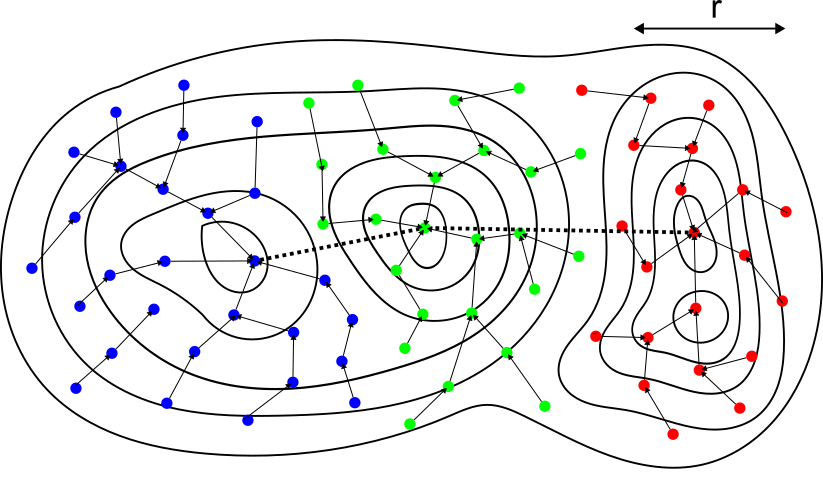}
  \caption{The construction of maximum transition network with a step-size $r$}
  \label{fig:nbn_with_r}
\end{figure}

NBN fundamentally represents the maximum transition network. This straightforward proof explains why NBN is effective: it simplifies the structure of the fitness landscape while still preserving its essential features. It is worth noting that NBN is not equivalent to the maximum transition network with the step-size $r$. As shown in Fig.\:\ref{fig:nbn_with_r}, when $\| \mathbf x, \beta(\mathbf x)\| > r$, the connection between two solutions is severed. 


\section{Calculation of the Nearest-better Network}\label{sec:NBN_calculation}
The first version of NBN was proposed in\:\cite{Diao2023Nearest}, where the NBN is generated by traversal algorithms (CNBSI),
in which the time complexity is O($N^2D$) ($N$ is the number of sampled solutions and $D$ is the dimensionality of the problem). Here, we provide a more efficient algorithm to calculate the network for the assignment problem, which represents a typical type of combinatorial problem\:\cite{Raggl2018discrete}, e.g., TSP and OneMax.

\subsection{Distance metric}
From the definition of NBN, 
the relationship between two solutions is defined based on the distance between two solutions, and the distance metric is different for different problems.

In One-Max problems with $D$ digits, the Hamming distance is used as the distance metric.
In the symmetric TSP with $D$ cities, the Dice coefficient\:\cite{dice1945measures} is used as the distance metric.
In the symmetric TSP, the distance between two solutions $\mathbf a$ and $\mathbf b$ defined by the Dice coefficient is:
\begin{equation}
  \| \mathbf a ,\mathbf b  \|  =  1 - 
  \frac{2 | \boldsymbol M(\mathbf a) \bigcap  \boldsymbol M(\mathbf b) |}{ |\boldsymbol M(\mathbf a) |+ | \boldsymbol M(\mathbf b) |} 
\end{equation}
where  for a solution $\mathbf a = [a_1,a_2,...,a_D]$, 
$\boldsymbol M(\mathbf a) = \left\{  (a_i, a_{(i+1)\%D}),(a_{(i+1)\%D}, a_i ) \mid   i = 1, 2, ...,D \right\}$
is the set of all edges that connect two cities for the solution. The neighborhood is defined according to the corresponding distance definition:
\begin{equation}
  \label{equ:dis_neighbor}
  \boldsymbol N(\mathbf x) = \left\{ \mathbf y \mid  \| \mathbf x , \mathbf y\| <r  , \forall \mathbf y\in \boldsymbol X_N \right\}
\end{equation}
where $r$ is a predefined value. 

\subsection{Problem representation}

We now consider the assignment problem\:\cite{Dorigo2019ant}, which widely exists in real-world applications. In the model, the search space is defined by a finite set of variables, $\mathbf x \in \mathbf V_1 \times \mathbf V_2 \times ,...,\times  \mathbf V_D $,
and each variable of the solution $x_i$ has an associated domain $\boldsymbol V_i$ of values that can be assigned to it.
A solution  $\mathbf x = [x_1,x_2,...,x_D]$ is an assignment of a value $v \in \boldsymbol V_i$ to variable $x_i$ and it is denoted by $x_i = v$.
In a TSP with $D$ cities, a variable $x_i$ in a solution 
is a move from city $i$ to the  city $x_i$.
This can be formalized by associating one variable to each city 
and each variable $x_i$ has then  $D-1$ associated values, $x_i \in \mathbf V_i = \{  a \mid a= 1,2,...,D, a \neq  i\}$.
In a OneMax problem with $D$ digits, a variable $x_i$ indicates the digit at the $i^{th}$ dimension, $x_i \in \mathbf V_i = \left\{ 0, 1 \right\}$ .





\subsection{The proposed algorithm}

\begin{algorithm}[!htbp] 
  \small
  \renewcommand{\algorithmicrequire}{\textbf{Input:}}
\renewcommand{\algorithmicensure}{\textbf{Output:}}
\caption{Calculation of nearest better solutions by division (CNBSD)} 
\label{alg:nbn_division} 
\begin{algorithmic}[1] 
    \REQUIRE  A set of sampled solutions  $\boldsymbol S$, \\
    the set of all the unselected variable domain set $\boldsymbol D_{\boldsymbol V}$,\\
    and the minimum number of the size of the calculated set $N_m$
    \ENSURE The nearest better relationship $ \beta$,
    and the best solution $\mathbf b$ of $\boldsymbol S$.
    \IF{$|\boldsymbol S| \leq N_m  $}
    \STATE $\beta = $CNBSI$ (\boldsymbol S)$
    \STATE $\mathbf b=  \arg\max_{\mathbf a \in \boldsymbol S } f(\mathbf a)$
    \ELSE 
    \STATE Record all the best solutions for each subset $\boldsymbol P  = \emptyset$
    \STATE Randomly select a domain set $\boldsymbol V_k$ from $\boldsymbol D_{\boldsymbol V}$
    \STATE $\boldsymbol D_{\boldsymbol V}= \boldsymbol D_{\boldsymbol V} - \left\{ \boldsymbol V_k\right\} $ 
    \STATE Divide $\boldsymbol S$ by the $k^{th}$ domain set $\boldsymbol V_k$ 
    into different subset, $\boldsymbol S = \boldsymbol S^1\cup...\cup\boldsymbol S^t  $
    \FOR{each subset $\boldsymbol S_{\mathbf x}^i$  } 
    \STATE $( \beta_i,\mathbf b_i ) = $CNBSD$(\boldsymbol S^i, D_{\boldsymbol V}, N_m)$
    \STATE $\beta = \min (\beta ,\beta_i)$
    \STATE $\boldsymbol P = \boldsymbol P \cup  \left\{  \mathbf b_i \right\}$
    \ENDFOR
    \STATE $\beta_b = $CNBSI$ (\boldsymbol P)$
    \STATE $\beta = \min (\beta ,\beta_b)$
    \STATE $\mathbf b=  \arg\max_{\mathbf a \in \boldsymbol P } f(\mathbf a)$
    \ENDIF
\end{algorithmic}
\end{algorithm}

Inspired by the random projection technique\:\cite{Bingham2001random}, 
which serves as an efficient dimensionality reduction technique for combinatorial problems, 
we propose a similar method to speed up the calculation.

In the proposed method, we divide the solution set into several smaller solution sets by one of the domains associated with one dimension, until the solution set is small enough to calculate their nearest better solutions, normally with $N_m= 20$ solutions. For one divided solution set, if it is divided into multiple subsets, 
it gathers the best solution from each subset to calculate its NBN by CNBSI; otherwise, the NBN is calculated directly by Algorithm\:\ref{alg:nbn_division}.

As stated in Line 8 of Algorithm\:\ref{alg:nbn_division}, a solution set \( \boldsymbol S \) is randomly divided into multiple subsets. and CNBSD is used to compute the best solution and nearest better relationships (\( \beta \)) for each subset. For the nearest better relationships of the current solution set  \( \boldsymbol S \), the nearest better relationships \( \beta \) for the solutions within each subset have already been calculated, as shown in Line 10.  For each subset's best solution, the nearest better solution (excluding the best solution of the current solution set) must belong to the current solution set, which must be the best solution of the other subsets. Therefore, we can use CNBSD to compute the nearest better relationships for the set of best solutions \( \boldsymbol P \) from these subsets, as shown in Line 14.

Finally, by merging the results of the nearest better relationships (i.e., select the closest better solution for each solution), we obtain the nearest better relationship \( \beta \) for the current solution set \( \boldsymbol S \). With more random partitionings, the accuracy of the nearest better relationship calculation is more accurate. This is done in Algorithm\:\ref{alg:CNBS-CAP}.


Actually, dividing the solution set into several subsets is a projection for the solutions 
from the original search space to a new specific dimension, and there exists an error in the calculation of the nearest neighbors with only one projection.
According to Johnson-Lindenstrauss lemma\:\cite{Kleinberg1997two}, we can calculate NBN with $N$ solutions with the minimum number of projections, $L$, and the desired error limit $\epsilon$, such that the solutions can be projected with a high probability based on a random projection:
\begin{equation}
  L  > \frac{\ln(N)}{\epsilon^2}
\end{equation} 
Thus, we can guarantee that the error of Algorithm.\:\ref{alg:CNBS-CAP} is smaller than $\epsilon$ with $L  > \frac{\ln(N)}{\epsilon^2} $ times of projections.

Generally, for a set of $N$ sampled solutions, in a single random projection, we only need to partition a solution set 
$N$ times to completely distinguish any two solutions. For each pair of solutions, we compute the distance between them to calculate NBN. Therefore, the time complexity of this algorithm is $\frac{N\ln(N)D}{\epsilon^2}$, where $D$ is the dimensionality of the problem, e.g., the number of cities in a TSP instance.
\begin{algorithm}[t] 
  \renewcommand{\algorithmicrequire}{\textbf{Input:}}
\renewcommand{\algorithmicensure}{\textbf{Output:}}
\caption{Calculation of the nearest better solutions by random projection (CNBSRP)} 
\label{alg:CNBS-CAP} 
\begin{algorithmic}[1] 
    \REQUIRE  A set of sampled solutions  $\boldsymbol S$,\\
     the number of loops for calculation $L$, \\
     and the minimum number of the size of the calculated set $N_m$
    \ENSURE The nearest better relationship $ beta$ 
    \FOR{ $k \leftarrow  1 $ to $L$} 
    \STATE Initialize $\boldsymbol D_{\boldsymbol V}$:  $\boldsymbol D_{\boldsymbol V} = \left\{\boldsymbol V_1, \boldsymbol V_2,..., \boldsymbol V_D  \right\} $ 
    \STATE $( \beta_i,\mathbf b_i ) = $CNBSD$(\boldsymbol S, D_{\boldsymbol V}, N_m)$
    \STATE Update $\beta$ :\: $\beta_i = \min (\beta, \beta_i  )$
    \ENDFOR
\end{algorithmic}
\end{algorithm}

\begin{figure}[!htbp]
  \centering
  \includegraphics[width=0.45\textwidth]{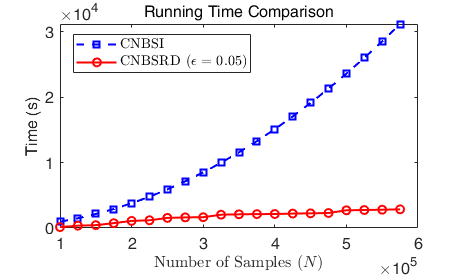} 
  \caption{Running time of the two algorithms, 
 where both algorithms are implemented using multithreading in a system equipped with an Intel(R) Xeon(R) CPU E5-2699 v4 @ 2.20GHz, featuring 88 cores.} 
  \label{fig:alg_time} 
\end{figure}

As shown in \text{Fig.\:\ref{fig:alg_time}}, compared to the previous NBN algorithm\:\cite{Diao2023Nearest} with a time complexity of $O(N^2 D)$, 
this algorithm has significantly reduced the time complexity. 
Moreover, this algorithm is parallelizable. In Algorithm.\:\ref{alg:CNBS-CAP}, we can parallelize the processing of each random projection, 
which is beneficial for optimizing the use of computing resources and for accelerating the NBN calculation.

\subsection{Local sampling}
The solution space of a combinatorial optimization problem typically consists of a large number of solutions. For example, 
in a TSP with $D$ cities, the number of solutions is $(D-1)!/2$. 
If we perform global sampling with only $N$ sampled solutions, 
the distribution of the sampled solutions is relatively sparse compared to the entire solution space, and we can only observe the global structure of the problem. The local structure of the problem is almost impossible to observe. 
To observe the problem from different scales, we can perform a local sampling in a local area around a center solution $\mathbf x_c$ within a distance of $K$.
The local area around a center solution $\mathbf x_c$ within a distance of $K$, $\boldsymbol S(\mathbf x_c, K)$,  is defined by:
\begin{equation}
  \label{eq:local_area}
  \boldsymbol S(\mathbf x_c , K )= \left\{\mathbf a \in \boldsymbol X \mid  \| \mathbf a, \mathbf x_c \| \leq  K \right\}
\end{equation}

The proposed algorithm can directly calculate the NBN structure for local region sampled solutions. 
However, in our experiments, we found out that our algorithms perform well when the sampled solutions are evenly distributed. 
But when the distribution of sampled solutions is concentrated, 
some subsets may contain many solutions and it is time-consuming to divide such subsets.
In a local sampled solutions set,  many solutions share the same values with the center solution $\mathbf x_c$, 
and thus we remove the subset that contains $\mathbf x_c$  and this is done after Line 8 of Algorithm\:\ref{alg:nbn_division}.


\section{NBN assisted analysis of combinatorial problems and algorithms}\label{sec:NBN_analysis}

\subsection{Analysis metric}
In our previous work \cite{Diao2024Nearest}, we introduced several metrics for evaluating landscape features based on NBN, 
including modality, BoA, ruggedness, and neutrality.
These metrics were originally proposed under the assumption of a uniformly distributed dataset. 
However, local or algorithmic sampling data is non-uniform, rendering the previously proposed metrics inapplicable to the biased data-based NBN scenarios.
In this subsection, we introduce some new metrics that do not rely on the assumption of uniform data distribution, thereby assisting in analyzing problems and algorithms.

\begin{itemize}
	\item  Evolutionary path ($\tilde{P}$)
	
	NBN, $\mathbf{G} = ( \boldsymbol V, \boldsymbol E) $, is essentially a tree structure, 
	where each solution is attracted by only one nearest better solution except for the global optima. For each solution $\mathbf{x} \in \boldsymbol V$, 
	there exists a path connecting $\mathbf{x}$ to the global optima $\mathbf{o}$.
	Considering that NBN is the maximum transition probability network, 
	it can be proven that this path represents the evolution path with the highest probability for the solution $\mathbf{x}$ to converge to the global optimum.
	The definition of this evolutionary path is defined as follows:
	
	$\tilde{P} (\mathbf{x}, \mathbf{o}) = [\mathbf{p}_1, \mathbf{p}_2, \ldots, \mathbf{p}_k]$
	where $\mathbf{p}_1 = \mathbf{x}$, $\mathbf{p}_k = \mathbf{o}$, and $k$ represents the number of nodes along the path.

  \item Distance of an evolutionary path ($d(\tilde{P})$)
  
  For any given evolutionary path \( \tilde{P} \), its complexity can be gauged by the maximum distance between its nodes. 
  An increased maximum distance implies a reduced likelihood of transitioning to the subsequent node, 
  thereby indicating a higher degree of difficulty of the evolutionary path. 
  This measure is quantified as the distance of the evolutionary path in this paper, defined as \( d(\tilde{P}) = \max_{i=1}^{k-1} \| \mathbf{p}_i, \mathbf{p}_{i+1} \| \).


  \item  Distance of a solution set to the optima ($d(\boldsymbol T, \mathbf{o})$)
  
  Evolutionary algorithms focus on solutions with either higher fitness values or greater evolutionary potential. 
  As long as there is one solution in the  solution set, \( \boldsymbol{T}\), that possesses a shorter evolutionary path, 
  the algorithm based on this set is more likely to converge to the global optimum.
  
  Accordingly, this paper defines the distance of the shortest evolutionary path of a solution set 
  as the distance between a solution set and the global optimum, denoted as:
  \begin{equation}
    d(\boldsymbol{T}, \mathbf{o}) = \min_{\mathbf t \in \boldsymbol{T}  }  d(\tilde{P}(\mathbf t) ) ,
    \end{equation}
    Furthermore, the shortest evolutionary path from the solution set to the global optimum is given by:
  \begin{equation}
    \tilde{P}(\boldsymbol T, \mathbf o) = \tilde{P}(\mathbf t), \mathbf t  = \arg \min_{\mathbf t \in \boldsymbol{T}  }  d(\tilde{P}(\mathbf t) )  
  \end{equation}

  \item Identification of optima in biased data-based NBN

In our previous work\:\cite{Diao2024Nearest}, 
optimal solutions are identified based solely on the magnitude of the Nearest-Better Distance (NBD) of the solutions. 
However, this approach is not suitable for the NBN generated from biased data. 
The distribution of data evolved by the algorithm is non-uniform. 
Early in the evolutionary process, the search radius of the algorithm is relatively large, 
leading to larger NBD values in some poorer regions. Consequently, some solutions may be mistakenly judged as local optima due to their large NBD.

Optimal solutions are inherently those with better fitness values. In the biased data-based NBN, 
fitness and NBD are integrated to identify optima, as shown in:
\begin{equation}
\label{eq:num_opt}
     f(\mathbf{x}) \geq \theta \, \land \, d_{\mathrm{NBD}} (\mathbf{x}) \geq \vartheta 
\end{equation}

\end{itemize}

\begin{figure*}[t]
  \centering

  \begin{tabular}{|c|c|c|c|}
  \hline
  origin &Ruggedness $\gamma = 4356$ & Neutrality $\mu = 24$ &  Epistasis $\upsilon = 14$ \\ 
  \includegraphics[width=0.225\textwidth]{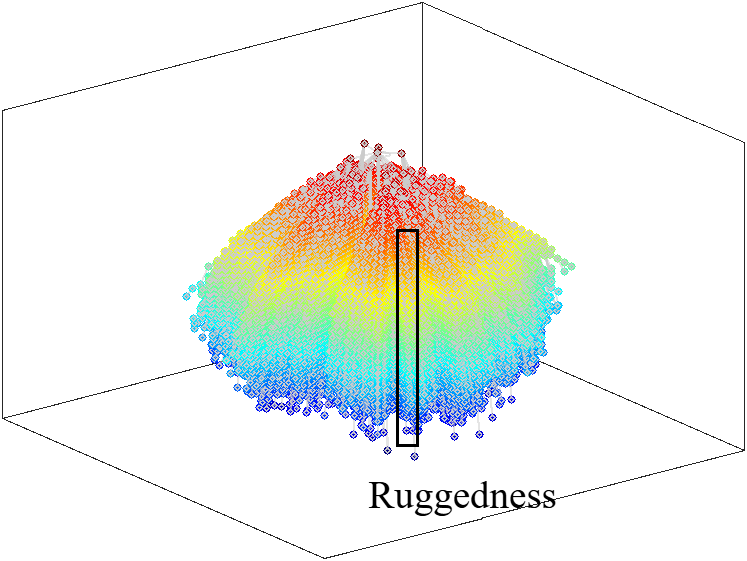}%
  &  
  \includegraphics[width=0.225\textwidth]{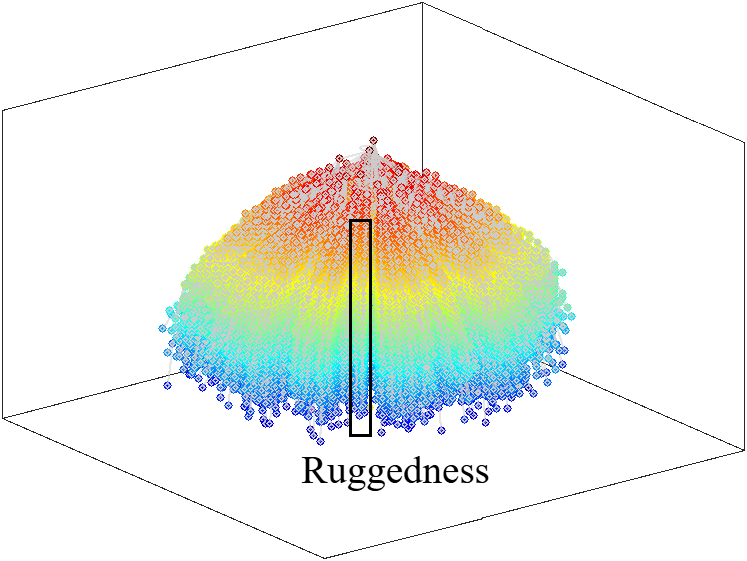}%
   &           
   \includegraphics[width=0.225\textwidth]{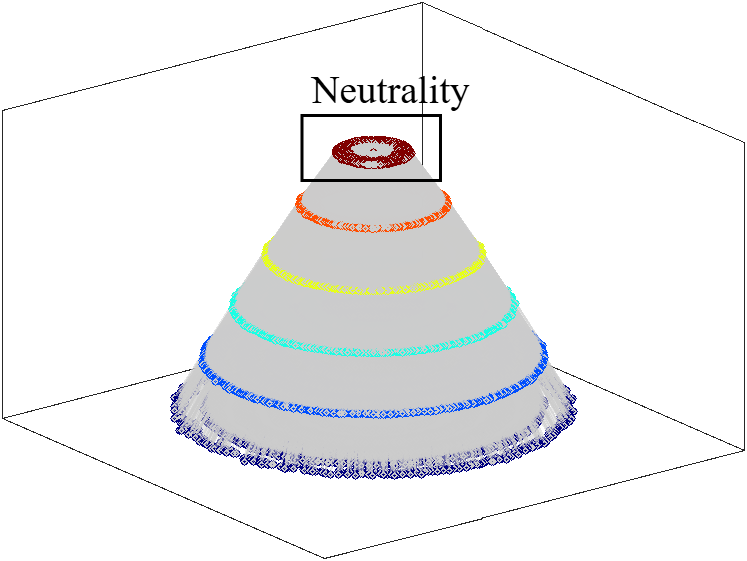}%
   &     
   \includegraphics[width=0.225\textwidth]{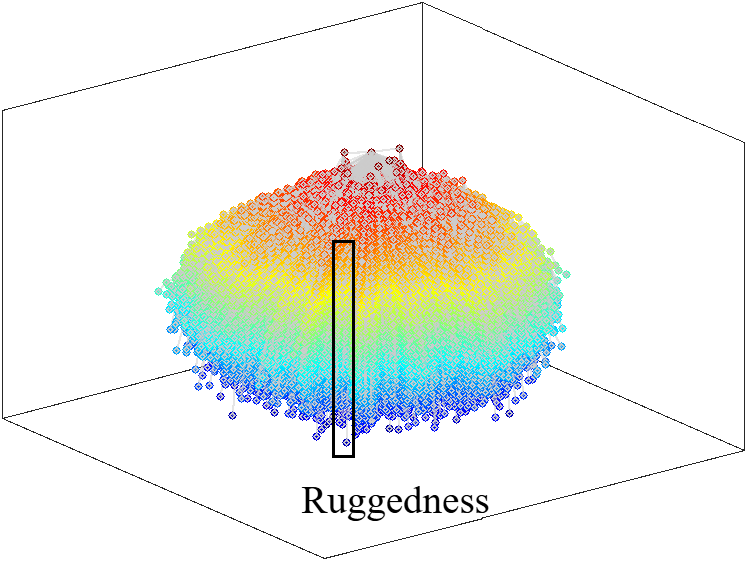}%

   \\
   $K = 120$ &   $K = 120$  &  $K = 120$&  $K = 120$ 
  \\

  \includegraphics[width=0.225\textwidth]{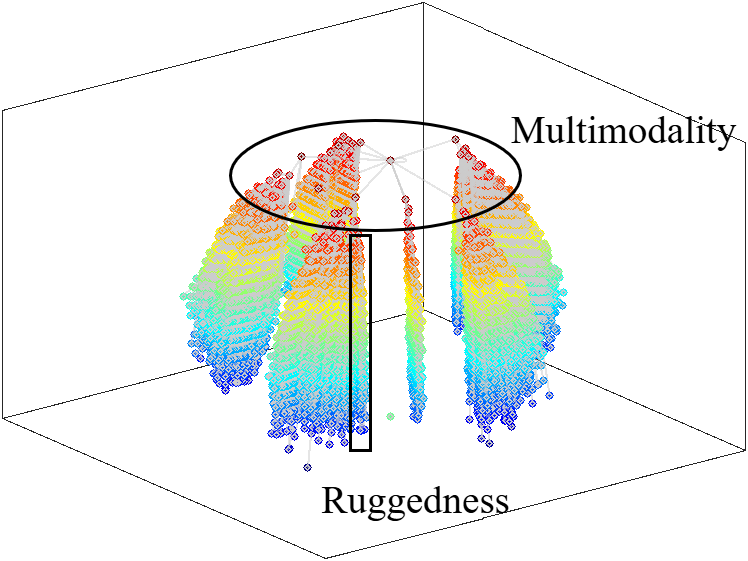}%
  &  
  \includegraphics[width=0.225\textwidth]{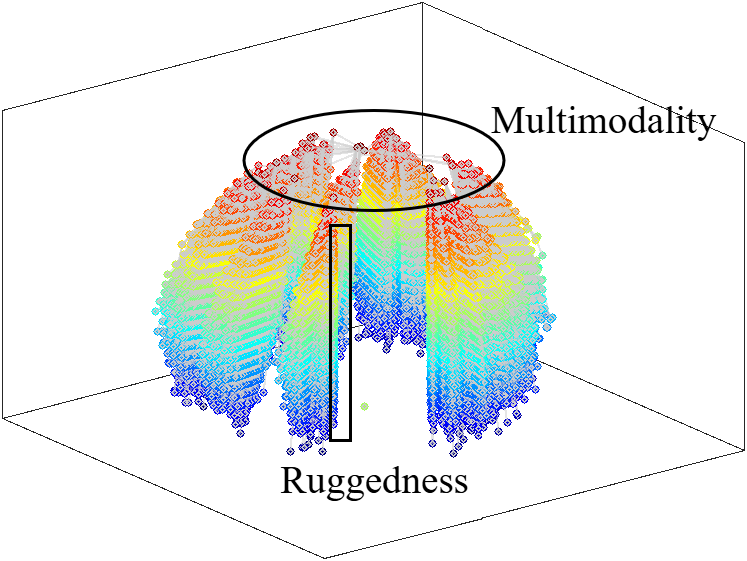}%
   &           
   \includegraphics[width=0.225\textwidth]{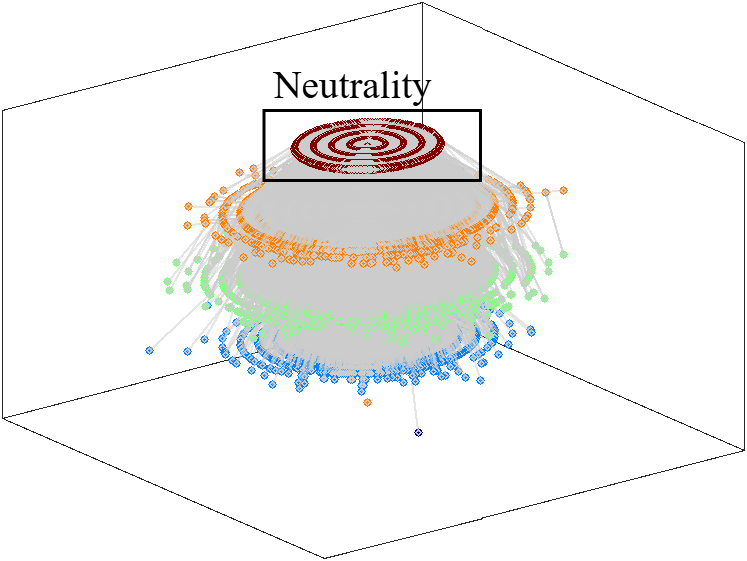}%
   &     
   \includegraphics[width=0.225\textwidth]{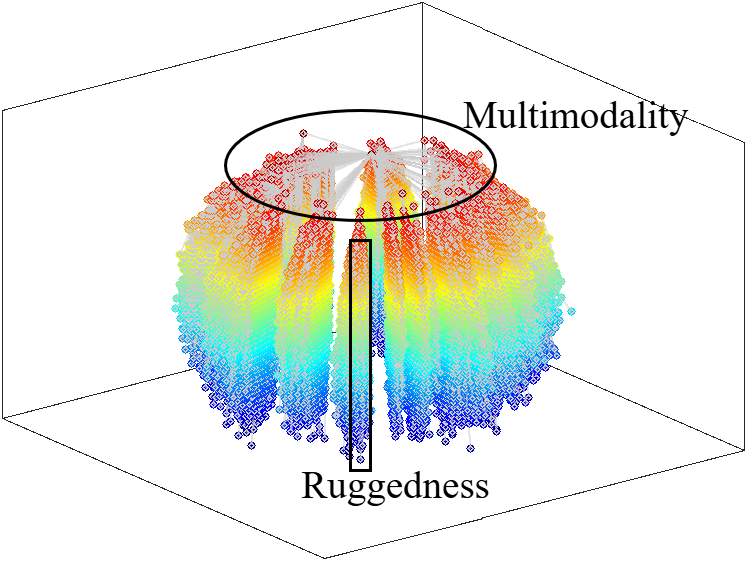}%
             
   \\
   $K = 30$ &   $K = 30$  &  $K = 30$&  $K = 30$ 
  \\

  \includegraphics[width=0.225\textwidth]{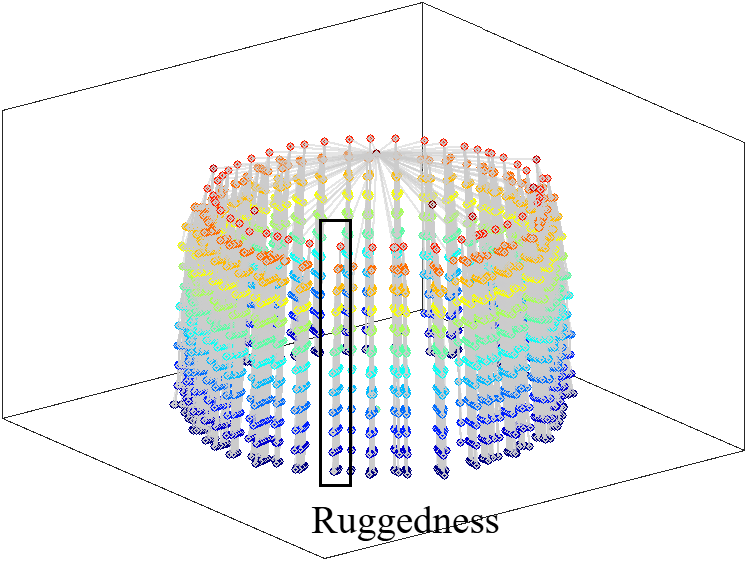}%
  &  
  \includegraphics[width=0.225\textwidth]{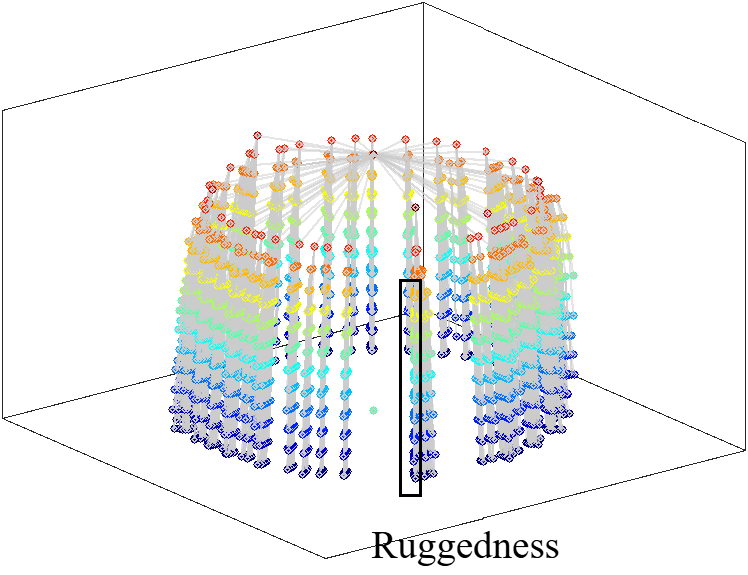}%
   &           
   \includegraphics[width=0.225\textwidth]{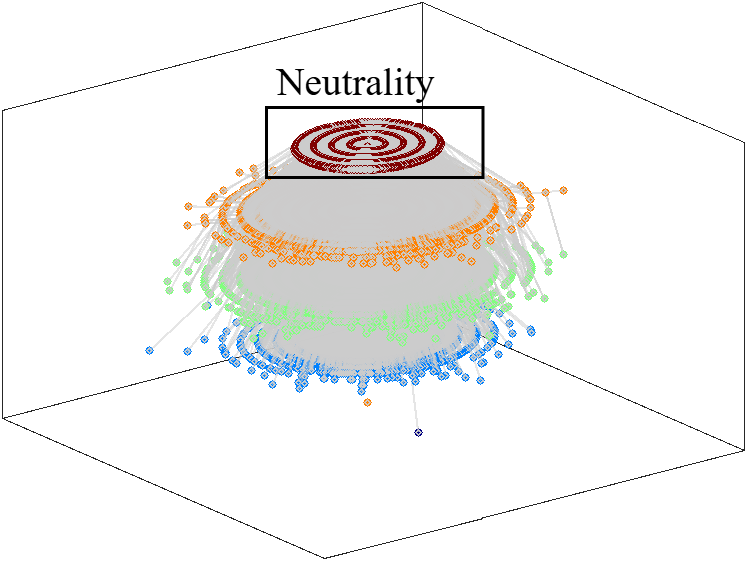}%
   &     
   \includegraphics[width=0.225\textwidth]{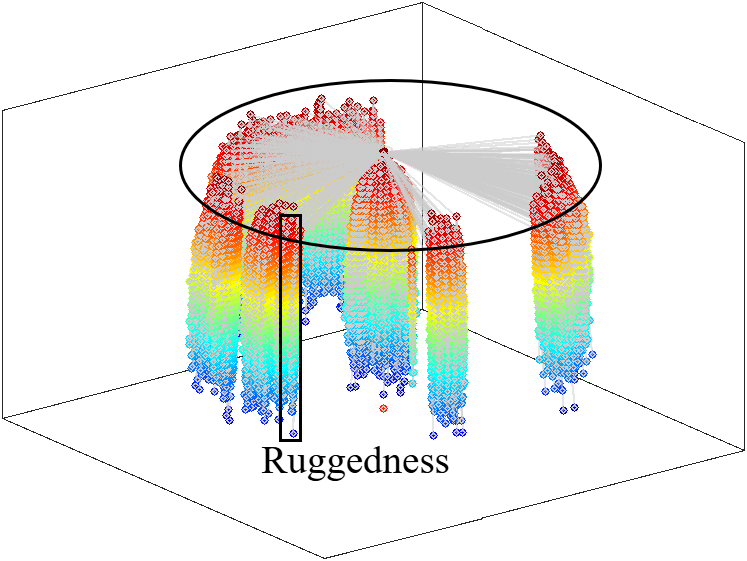}%
              
   \\
   $K = 7$ &   $K = 7$ &  $K = 7$&  $K = 7$ 

  \\ \hline

  \end{tabular}
    \caption{ Visualization of NBN on 120 bits Tunable W-Model  of different features from different scales }
  \label{fig:nbn_onemax_diff_scale}
  \end{figure*}

\subsection{Feature analysis of the OneMax problem}
The W-Model\:\cite{Weise2018difficult} is the only combinatorial benchmark that can
adjust the degrees of ruggedness, neutrality, and epistasis.
By setting specific parameters, we can modify the degree of the three features of a W-Model  function. The experiments in~\cite{Weise2018difficult} showed that these parameters indeed affect the problem-solving difficulty. However, there is currently no research proving whether these parameters influence difficulty by altering the corresponding landscape features or by changing other features. This subsection constructs W-Model  functions with different parameters and analyzes the global and local structures of the problem to analyze the correlation between these parameters and these features.

In this experiment, we construct W-Model  functions with 120 digits with different parameters,  where $\gamma$, $\mu$, and $\upsilon$ represent the degree of ruggedness, neutrality, and epistasis, respectively. Fig.\:\ref{fig:nbn_onemax_diff_scale} shows the impact of different parameters
on the global and local structures of the fitness landscape of W-Model  functions. TABLE~\ref{tb:onemax_num_opt} shows that the number of optima is quite different on different W-Model  functions based on Eq.\eqref{eq:num_opt} with $\theta = 9$ and $\vartheta =20$. 
In the experiments, the number of samples is set to $N = 10^6$, and $K$ is the radius of the local area for sampling as Eq.\:\eqref{eq:local_area}.

\begin{table}[!ht]
\caption{Identification of the number of optima for W-Model  of different features from different scales}
\label{tb:onemax_num_opt}
\centering
\begin{tabular}{|c|c|c|c|c|c|}
\hline
\multicolumn{6}{|c|}{Ruggedness}\\ \hline
\textbf{$\gamma$\textbackslash{} $K$}   & \textbf{7} & \textbf{15} & \textbf{30} & \textbf{60} & \textbf{120} \\ \hline 
\textbf{0}                              & 73         & 12          & 10          & 36          & 20           \\ \hline
\textbf{1452}                           & 65         & 12          & 16          & 21          & 19           \\ \hline
\textbf{2904}                           & 59         & 15          & 19          & 29          & 27           \\ \hline
\textbf{4356}                           & 60         & 9           & 18          & 29          & 18           \\ \hline
\textbf{5808}                           & 86         & 8           & 15          & 40          & 25           \\ \hline
\textbf{7260}                           & 69         & 11          & 12          & 23          & 24           \\ \hline
\multicolumn{6}{|c|}{Neutrality}\\ \hline
\textbf{$\mu$\textbackslash{} $K$}      & \textbf{7} & \textbf{15} & \textbf{30} & \textbf{60} & \textbf{120} \\ \hline 
\textbf{0}                              & 73         & 12          & 10          & 36          & 20           \\ \hline
\textbf{12}                             & 451061     & 258715      & 91916       & 22476       & 7440         \\ \hline
\textbf{24}                             & 453886     & 323253      & 194666      & 105314      & 65503        \\ \hline
\textbf{36}                             & 942718     & 809168      & 565516      & 336946      & 180960       \\ \hline
\textbf{48}                             & 994114     & 957815      & 831526      & 612520      & 309765       \\ \hline
\textbf{60}                             & 988072     & 932624      & 774689      & 542237      & 303673       \\ \hline 
\multicolumn{6}{|c|}{Epistasis}\\ \hline
\textbf{$\upsilon$\textbackslash{} $K$} & \textbf{7} & \textbf{15} & \textbf{30} & \textbf{60} & \textbf{120} \\ \hline
\textbf{0}                              & 73         & 12          & 10          & 36          & 20           \\ \hline
\textbf{2}                              & 34         & 14          & 11          & 27          & 57           \\ \hline
\textbf{6}                              & 98         & 34          & 52          & 114         & 165          \\ \hline
\textbf{10}                             & 401        & 89          & 110         & 154         & 179          \\ \hline
\textbf{14}                             & 508        & 95          & 125         & 173         & 178          \\ \hline
\textbf{18}                             & 397        & 66          & 107         & 150         & 201          \\ \hline
\textbf{22}                             & 922        & 148         & 134         & 168         & 165          \\ \hline
\end{tabular}
\end{table}

We analyze these W-Model  functions based on the following aspects:

\begin{itemize}
    \item Ruggedness 

As depicted in Fig.\:\ref{fig:nbn_onemax_diff_scale}, the NBN of the local structure of all W-Model  functions, except for the neutral functions, exhibit a rugged landscape, with straight hanging branches at the bottom\:\cite{Diao2024Nearest}. This aligns with the inherent understanding that the fitness landscapes of combinatorial problems are rugged and also explains why local search operators are crucial for these problems.

However, no remarkable differences are observed in the NBN visualization between the W-Model  functions with $\gamma = 0$ and those
with $\gamma= 4356$. Furthermore, \text{TABLE\:\ref{tb:onemax_num_opt}}
indicates that the parameter influences the number of optima, but it does not show a significant correlation between $\gamma$ and the number of optima.

\item Neutrality

From NBN of the neutral W-Model  functions with $\mu = 24$ in Fig.\:\ref{fig:nbn_onemax_diff_scale}, 
We can see several flat regions, indicating the function's neutrality feature. 
Moreover, \text{TABLE\:\ref{tb:onemax_num_opt}} shows that 
with a larger $\mu$, the function has a larger number of optima with the same fitness.
This shows that the parameter $\mu$ indeed affects the degree of neutrality of the functions. 

\item Epistasis

\begin{figure*}[t]
  \centering
  \begin{tabular}{|c|c|c|c|}
  \hline
  \multicolumn{4}{|c|}{rue500-1}   \\  \hline
  $K = 500$ &   $K = 200$  &  $K = 50$&  $K = 12$ \\

  \includegraphics[width=0.225\textwidth]{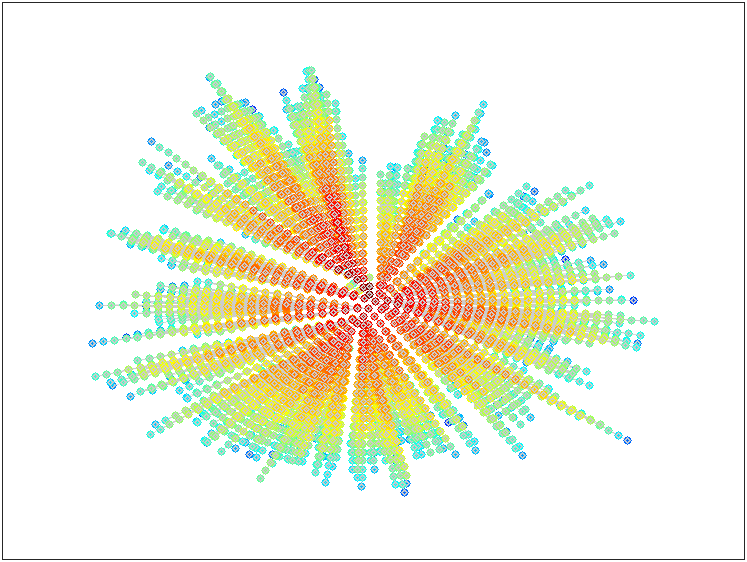}%
  &  
  \includegraphics[width=0.225\textwidth]{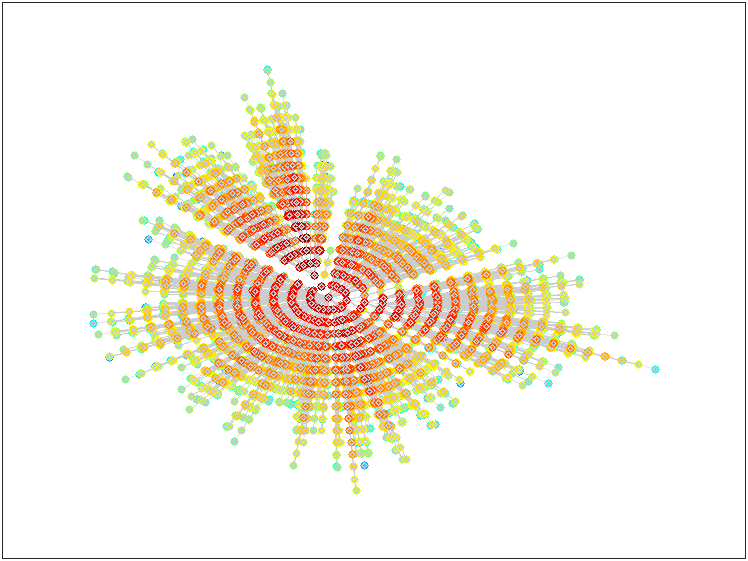}%
   &   
   \includegraphics[width=0.225\textwidth]{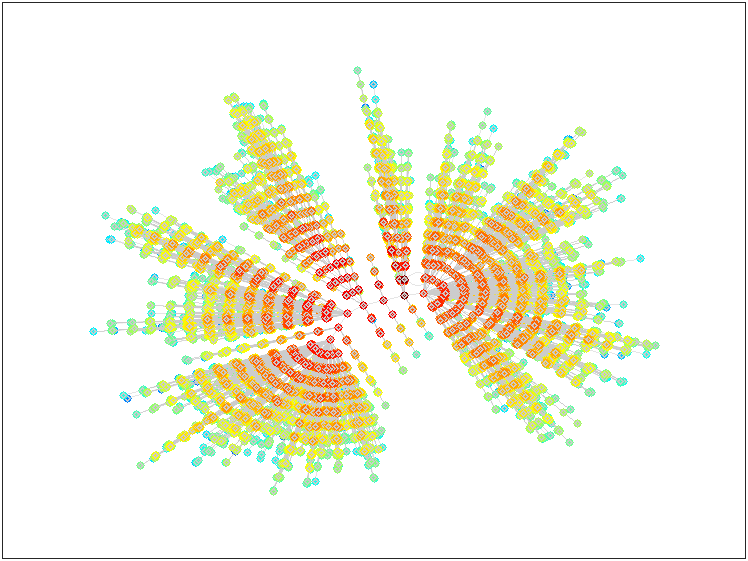}%
   &     
   \includegraphics[width=0.225\textwidth]{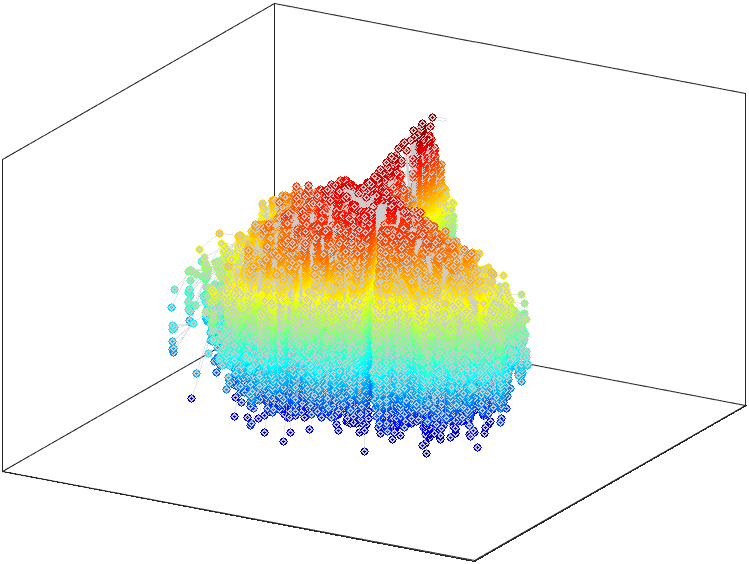}%
   \\
   top view &   top view &  top view&  side view 
  \\
  \includegraphics[width=0.225\textwidth]{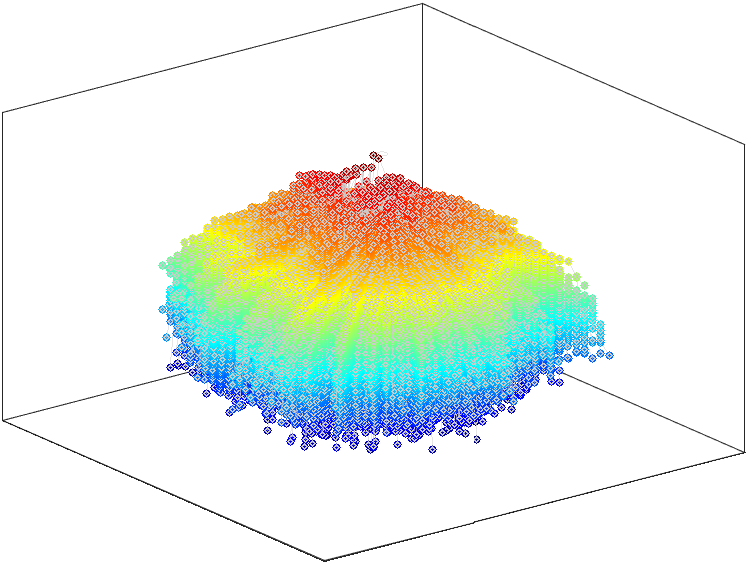}%
  &  
  \includegraphics[width=0.225\textwidth]{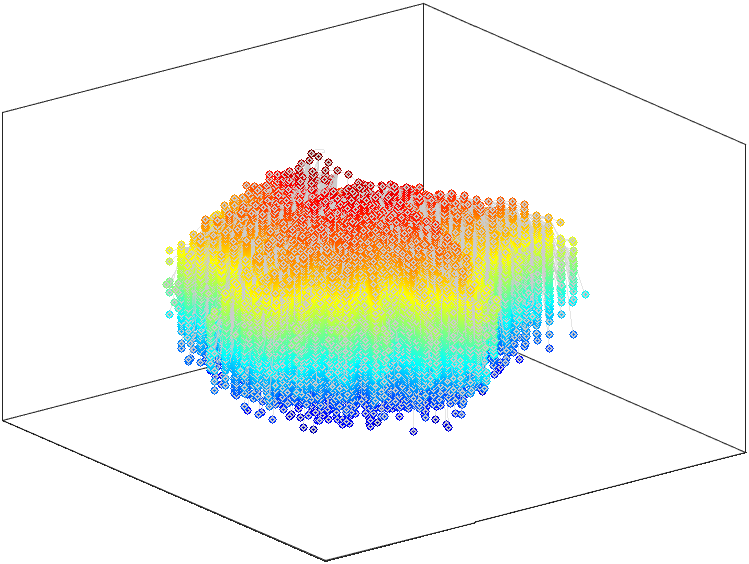}%
   &   
   \includegraphics[width=0.225\textwidth]{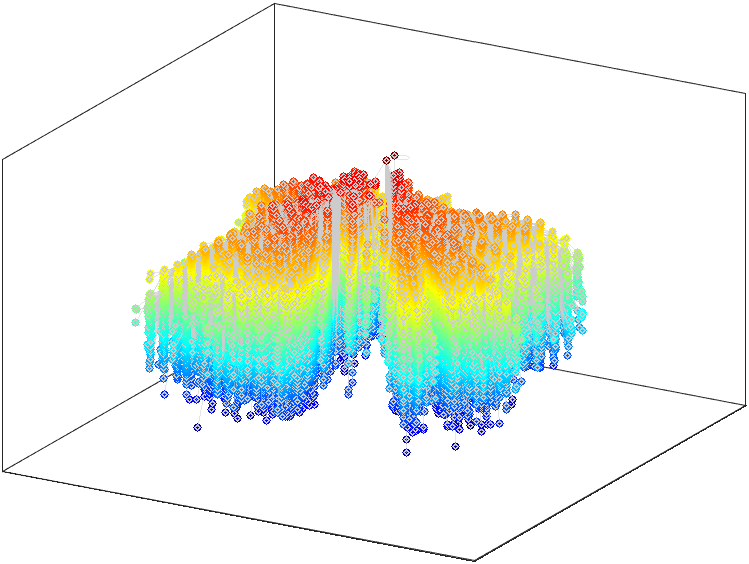}%
   &     
   \includegraphics[width=0.225\textwidth]{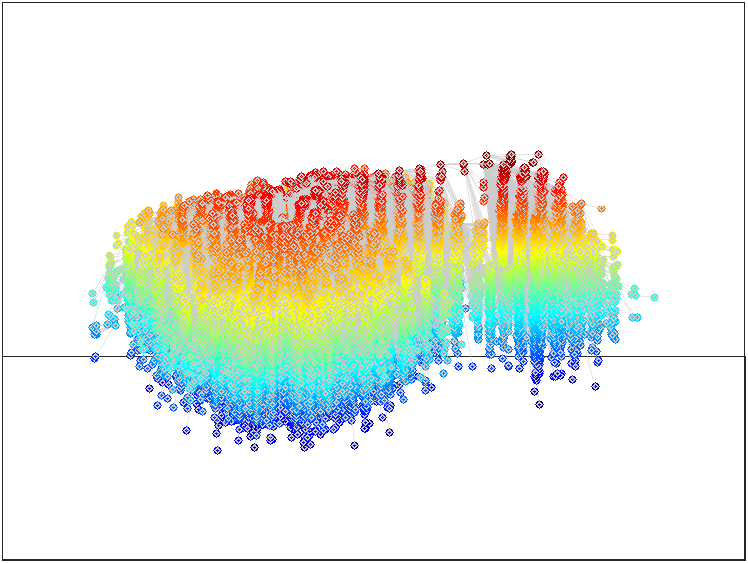}%
   \\
   side view &   side view &  side view&  side view 
  \\ \hline
  \end{tabular}
  \caption{Visualization of NBN of rue500-1 from different scales with $N = 10^6$ solutions generated by local random sampling, $K$ is the radius of the local region.}
  \label{fig:nbn_tsp_substructure}
  \end{figure*}

From Fig.\:\ref{fig:nbn_onemax_diff_scale}, we can see that the parameter $\upsilon$ essentially affects the number of optima and the size of BoAs, especially in the local structure of $K = 7$.  Furthermore, TABLE\:\ref{tb:onemax_num_opt} corroborates that there is a positive correlation between the $\upsilon$ and the number of optima. This indicates that the parameter indeed affects the degree of epistasis of the functions.

\end{itemize}

\subsection{Feature analysis of the traveling salesman problem}

This subsection aims to analyze the global and local structures of the fitness landscapes of three typical TSP instances and the search data of LKH and EAX. By combining this with the landscape features, this paper aims to identify the challenges for the algorithms in solving these problems. With the analysis, this paper aims to provide insights for the design and optimization of algorithms.

\subsubsection{TSP instances selection}

TSPlib~\cite{Reinelt1991TSP} is a widely used benchmark dataset and the portgen generator~\cite{Gutin2006traveling} generates TSP instances (referred to as a run instance) by randomly placing points on a two-dimensional plane.

We select three typical TSP instances: u574, rue500-1, and rue500-2. u574 is a commonly used TSP instance from TSPlib, containing 574 nodes. This instance was also used for observation and analysis in the paper of LON~\cite{Ochoa2018Mapping}. 
Researchers can compare the results of this paper with those of LON~\cite{Ochoa2018Mapping} to deepen the understanding of the fitness landscape of the TSP instances.
Both rue500-1 and rue500-2 have 500 nodes generated by the portgen generator. \text{TABLE~\ref{tb:eax_lkh_success}} shows the success rates of EAX and LKH, i.e., the number of runs that found the global optimum versus the total number of runs, where both algorithms use the recommended parameters.
As shown in TABLE~\ref{tb:eax_lkh_success}, the performance of LKH and EAX on the two TSP instances is the opposite. 
They both have 500 nodes, why do EAX and LKH behave so differently?
The analysis of this subsection tries to answer this question.


  \begin{table}[htp]
  \centering
  \caption{Success rate of EAX and LKH on different TSP instances}
  \label{tb:eax_lkh_success}
  \begin{tabular}{|c|c|c|c|}
  \hline
      & u574  & rue500-1 & rue500-2 \\ \hline
  EAX & 30/30 & 2/30     & 30/30    \\ \hline
  LKH & 30/30 & 30/30    & 4/30     \\ \hline
  \end{tabular}
  \end{table}
To answer the question above, we first take a look at the working mechanism of the two algorithms. LKH mainly consists of three techniques: local search operators, restart, crossover, and \textcolor{version_3}{a GA framework}. LKH will restart several times and in each restart, it generates a random solution, optimizes this solution using local search operators, and crosses over this solution with the iteration-best solution or \textcolor{version_3}{a solution in the GA population} to find a better solution.  The key to EAX lies in its GA framework and an efficient crossover operator. It is a single-population algorithm that maintains a certain level of diversity during the evolutionary process. 
  \begin{figure*}[!ht]
  \centering
    \begin{tabular}{|c|c|c|c|}
    \hline
    \multicolumn{4}{|c|}{u574}                                                                       \\ \hline
    LON     & NBN, LON data     & NBN, total data     &  \makecell{LDEE, EAX's success, $\boldsymbol{T}_0$}                \\ \hline
    \includegraphics[width=0.225\textwidth]{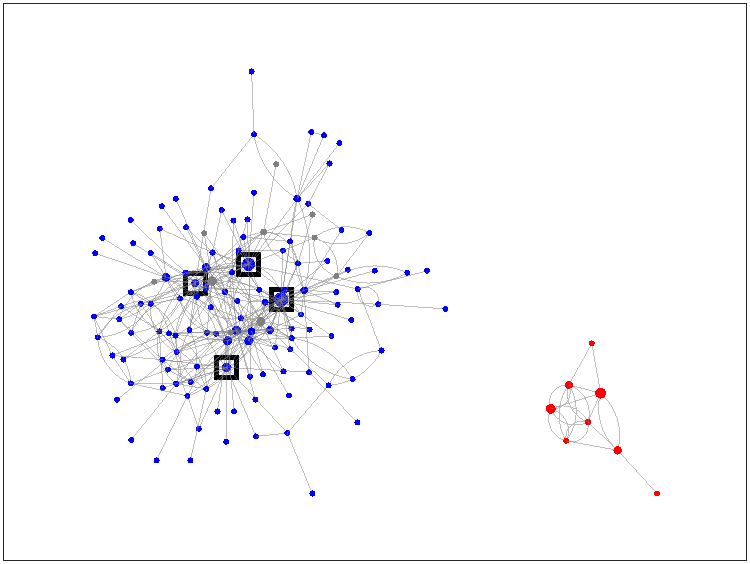}            
    & \includegraphics[width=0.225\textwidth]{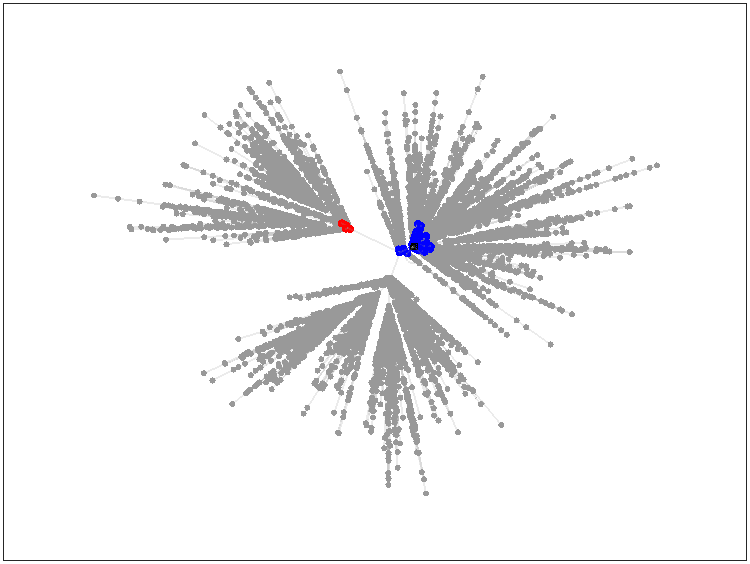}
    & \includegraphics[width=0.225\textwidth]{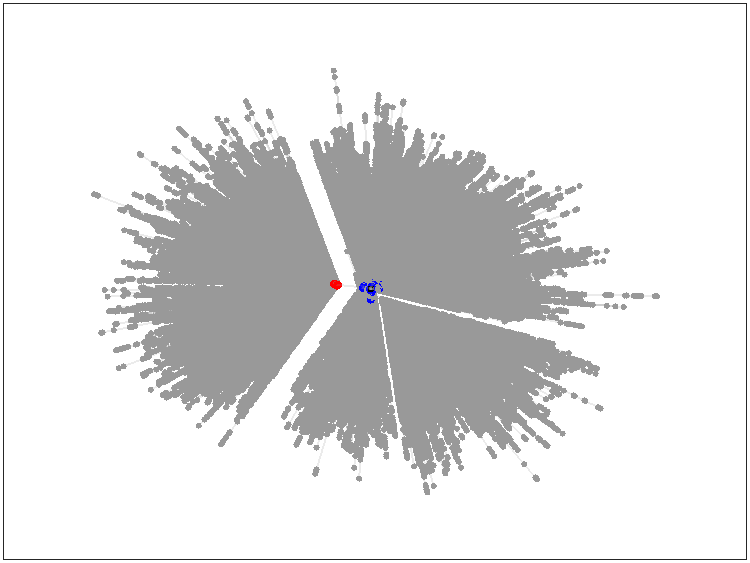}                       
    & \includegraphics[width=0.225\textwidth]{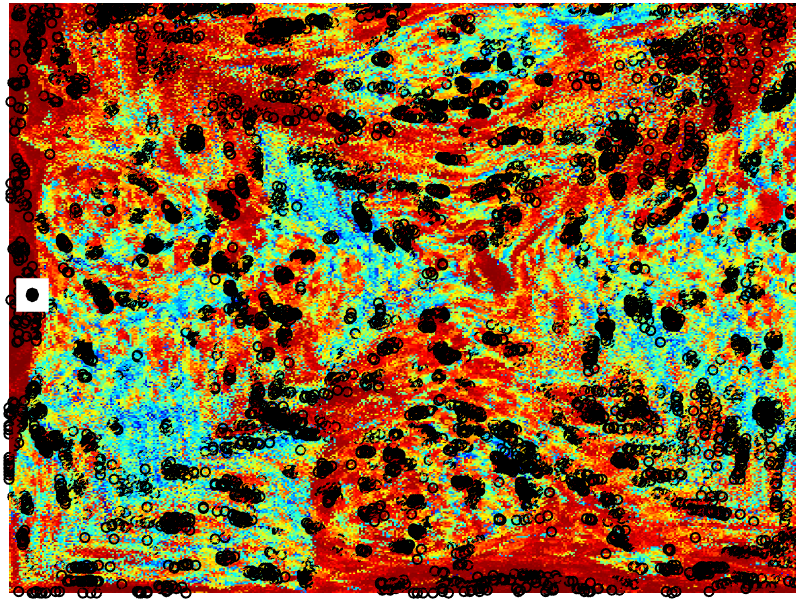}                         \\
    top view     & top view               & top view                & total data       \\
    \includegraphics[width=0.225\textwidth]{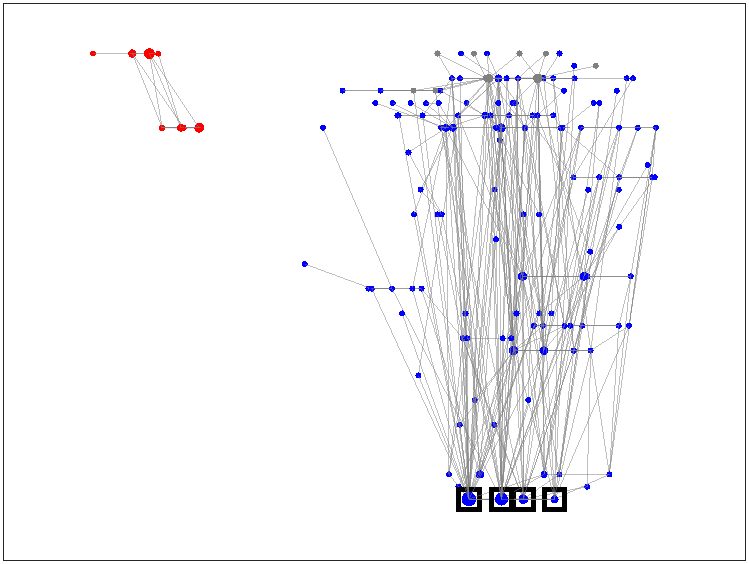}            
    & \includegraphics[width=0.225\textwidth]{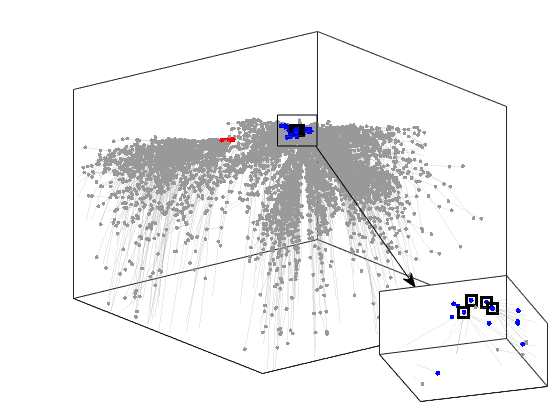}
    & \includegraphics[width=0.225\textwidth]{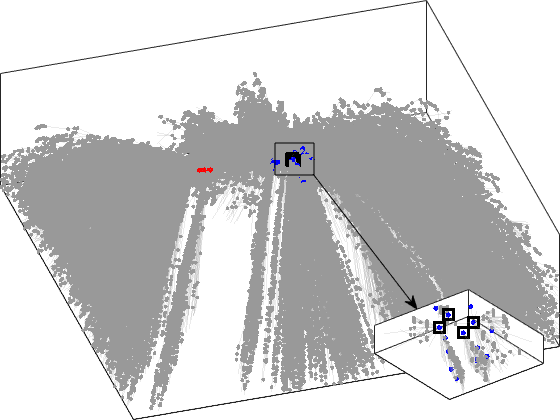}                       
    & \includegraphics[width=0.225\textwidth]{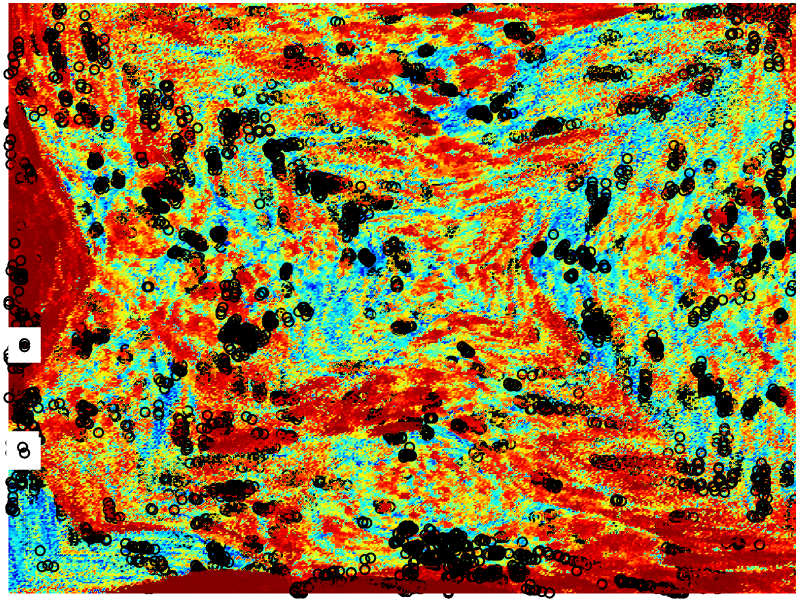}                         \\
    side view    & side view              & side view               & LON and EAX data \\ \hline
    \multicolumn{4}{|c|}{rue500-1}                                                                       \\ \hline
    LON & NBN, LON data & NBN, total data & LDEE, total data            \\ \hline
    \includegraphics[width=0.225\textwidth]{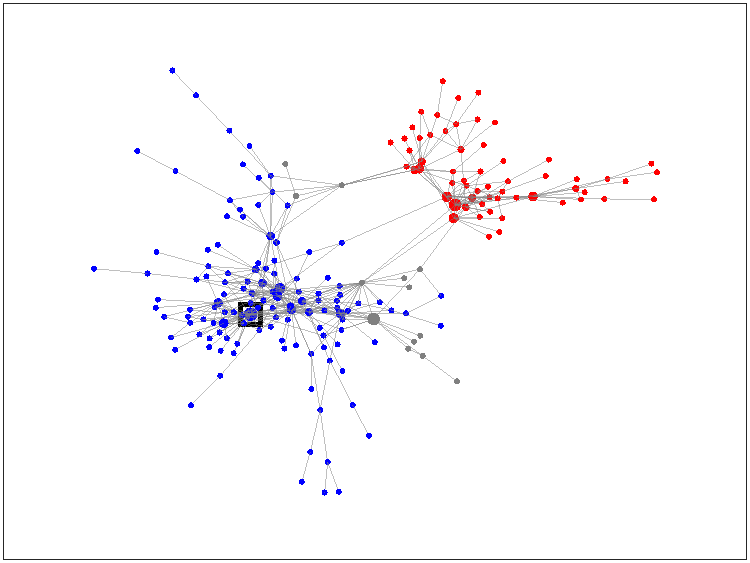}            
    & \includegraphics[width=0.225\textwidth]{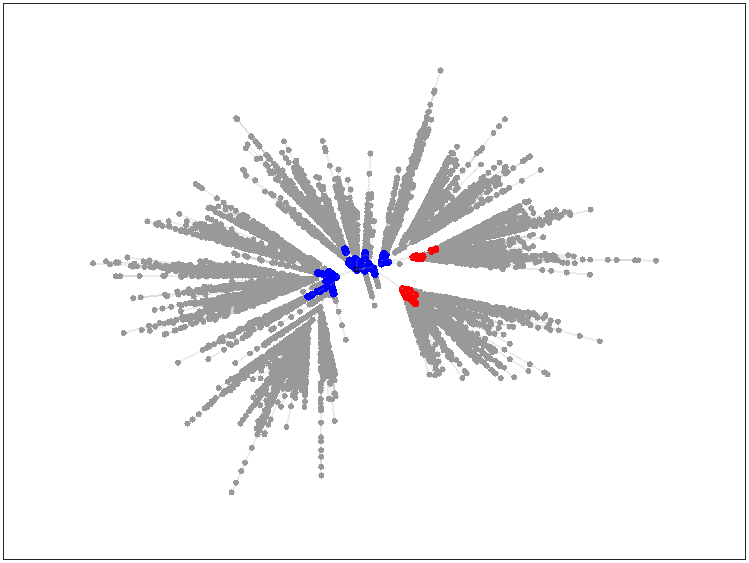}
    & \includegraphics[width=0.225\textwidth]{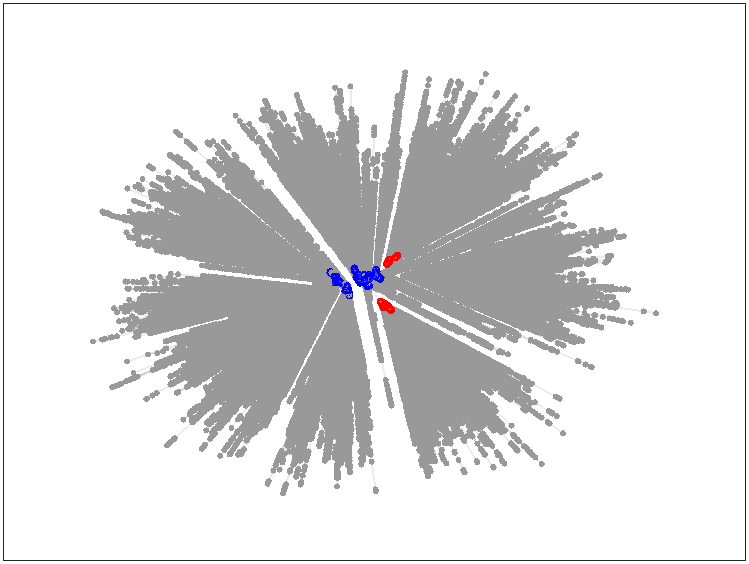}                       
    & \includegraphics[width=0.225\textwidth]{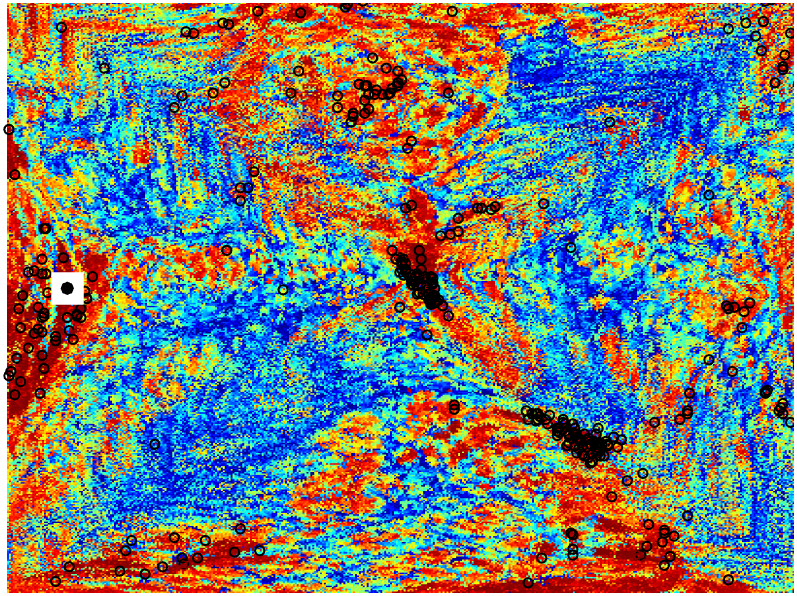}                         \\
    top view     & top view               & top view                & LKH 's success,  $\boldsymbol{T}_0$   \\
    \includegraphics[width=0.225\textwidth]{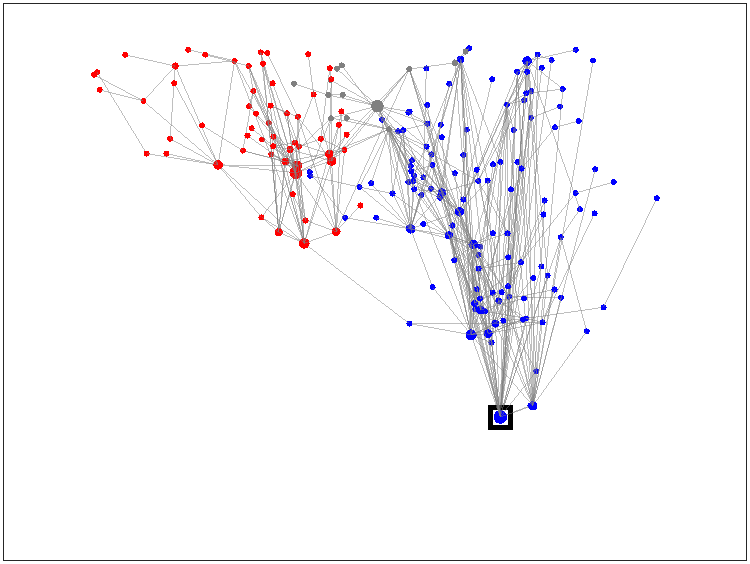}            
    & \includegraphics[width=0.225\textwidth]{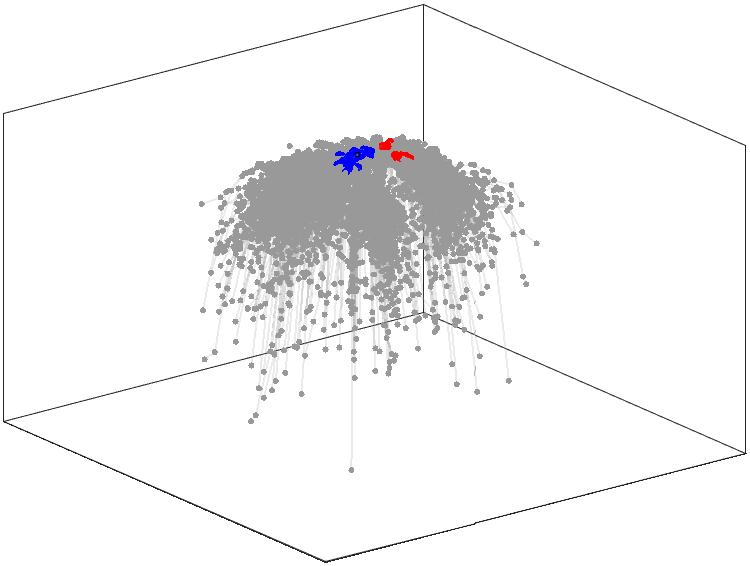}
    & \includegraphics[width=0.225\textwidth]{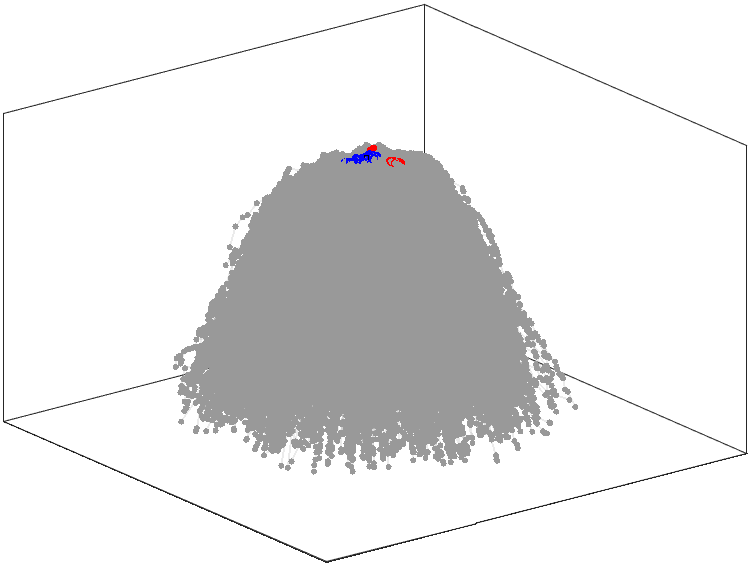}                       
    & \includegraphics[width=0.225\textwidth]{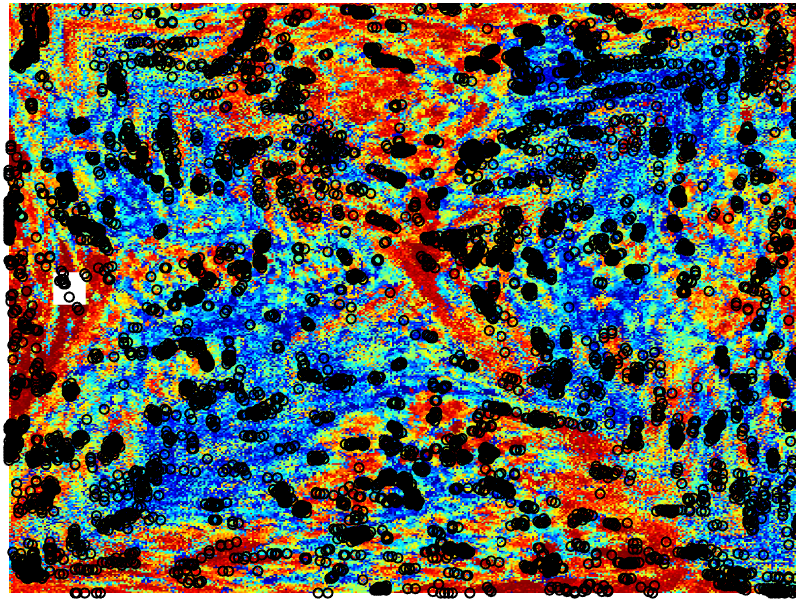}                         \\
    side view    & side view              & side view               & EAX 's failure,  $\boldsymbol{T}_9$   \\ \hline

    \end{tabular}

    \caption{Comparison of NBN, LON, and LDEE on u54 and rue500-1:  
        Recommended parameters\:\cite{Ochoa2018Mapping} are used in LON and figures of LDEE are generated by the tool provided by its authors\:\cite{Michalak2019Low}. There are three sources of data: data generated by LON, EAX, and LKH, that is LON data, EAX data, and LKH data. Total data is the union of all these data. 
     Each point in LON represents a local optimum, and connections between points indicate transitions between the solutions by the 4-opt operator. The height of each point indicates its fitness value (lower values indicate better fitness). The color of points (from blue to red) represents the basin to which they belong, while gray points belong to multiple basins. Black rectangles are the global optima. 
     This color-coding scheme is the same for NBN. 
     LDEE is a two-dimensional grid image in which each grid represents a solution. The color gradient from blue to red indicates the fitness value of solutions (red is the best). 
          White rectangles denote the locations of the optima, and black circles are the solutions of an algorithm's trajectory $\boldsymbol T$. $\boldsymbol T_i$ indicates the $i^{th}$ trajectory of an algorithm.}
    \label{fig:nbn_lon_compare1}
    \end{figure*}

\subsubsection{Global and local structure of TSP}
\begin{figure*}[!ht]
  \centering
    \begin{tabular}{|c|c|c|c|}
    \hline
    \multicolumn{4}{|c|}{rue500-2}  \\ \hline
    LON & NBN, LON data & NBN, total data & LDEE, total data             \\ \hline
    \includegraphics[width=0.225\textwidth]{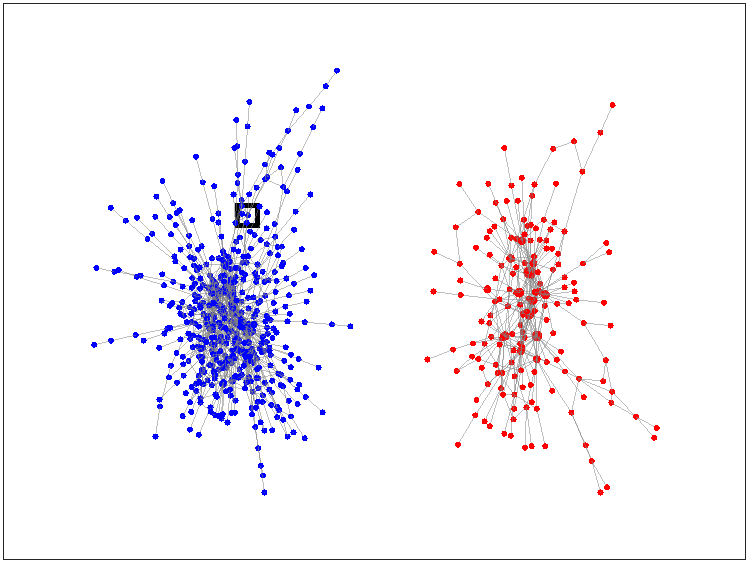}            
    & \includegraphics[width=0.225\textwidth]{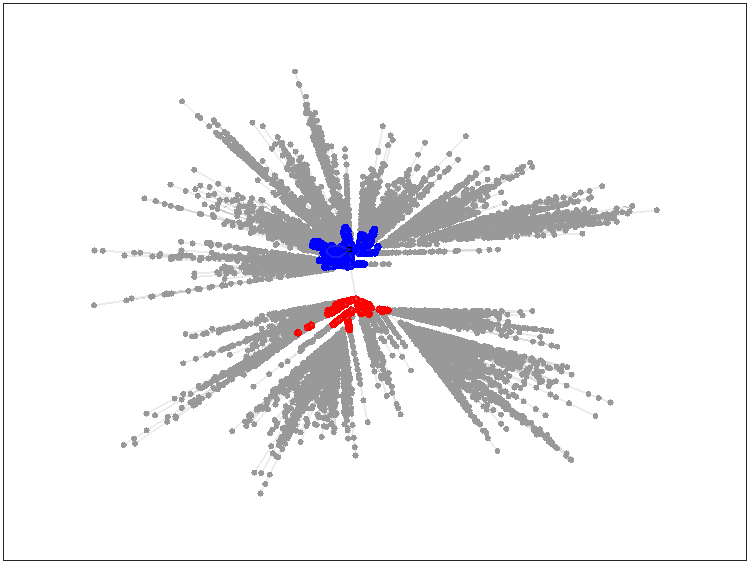}
    & \includegraphics[width=0.225\textwidth]{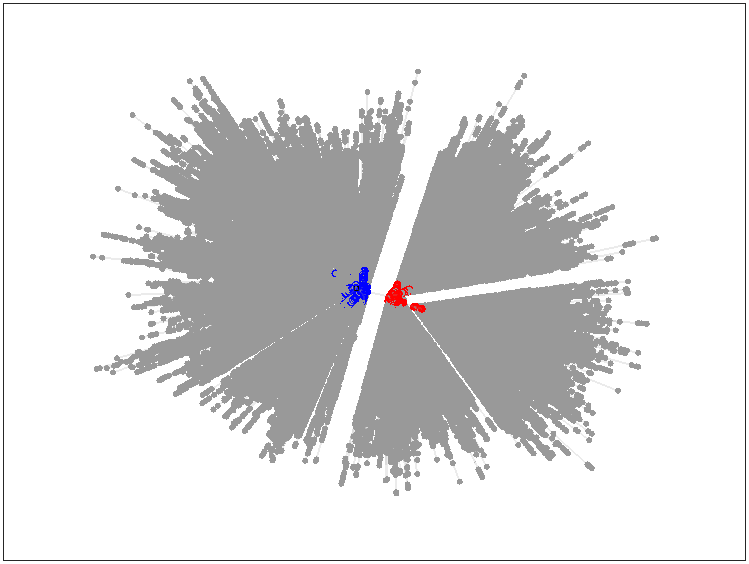}                       
    & \includegraphics[width=0.225\textwidth]{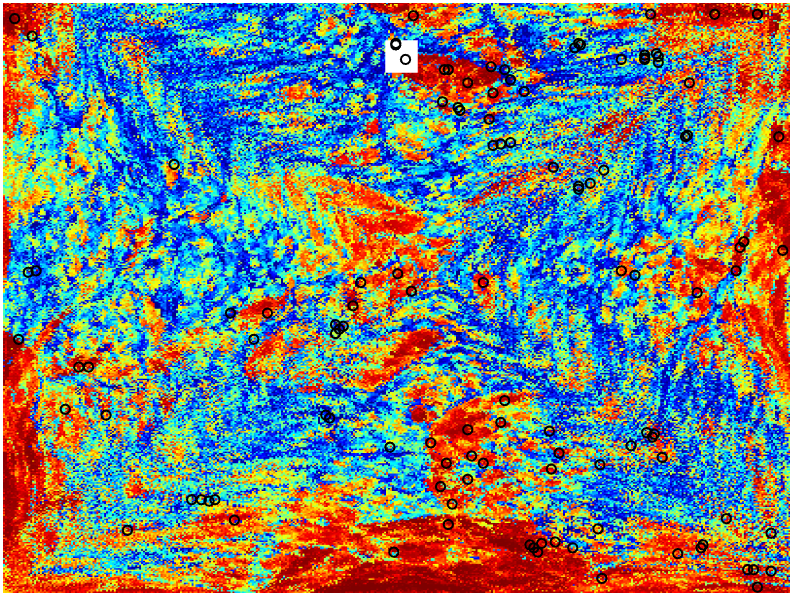}                         \\
    top view     & top view               & top view                & LKH 's failure,  $\boldsymbol{T}_0$  \\
    \includegraphics[width=0.225\textwidth]{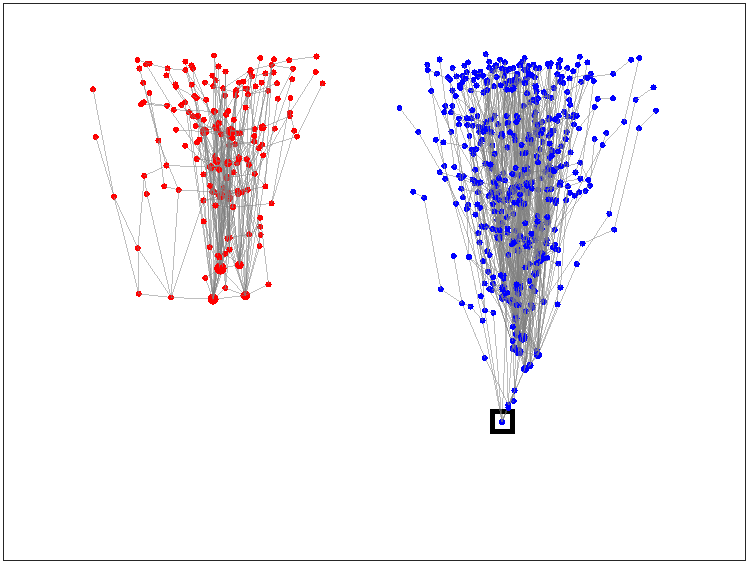}            
    & \includegraphics[width=0.225\textwidth]{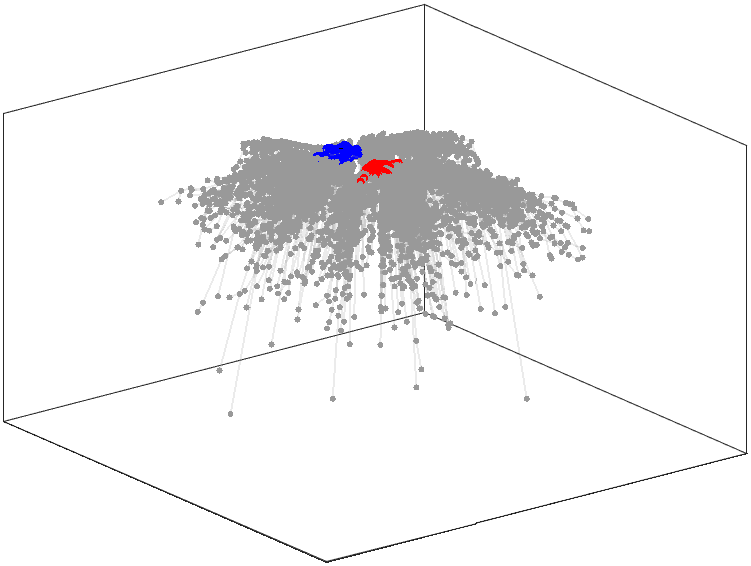}
    & \includegraphics[width=0.225\textwidth]{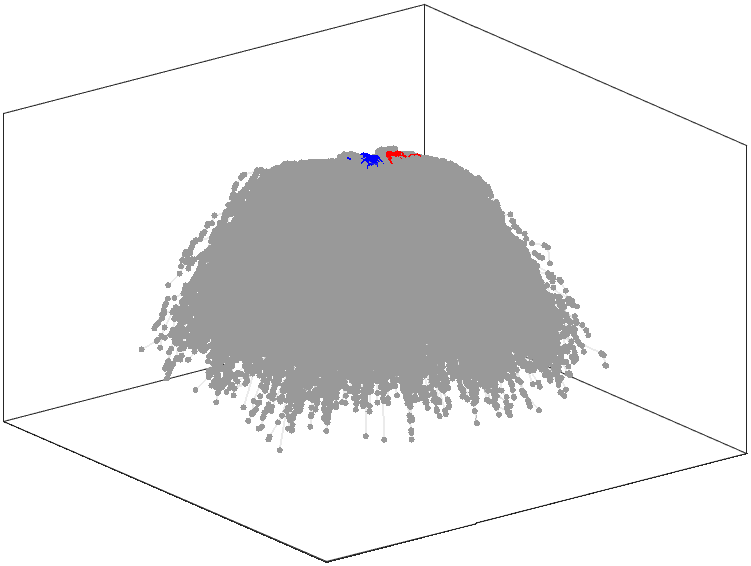}                       
    & \includegraphics[width=0.225\textwidth]{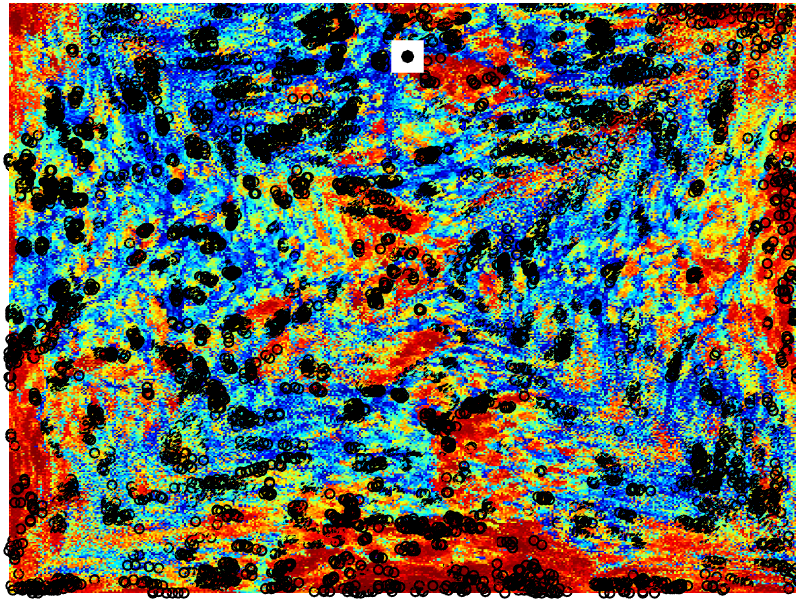}                         \\
    side view    & side view              & side view               & EAX 's success,  $\boldsymbol{T}_0$   \\ \hline
    \end{tabular}

    \caption{Comparison of NBN, LON, and LDEE on rue500-2}
    \label{fig:nbn_lon_compare2}
    \end{figure*}

By comparing the results of the three TSP instances in Fig.~\ref{fig:nbn_tsp_substructure}, Fig.~\ref{fig:nbn_lon_compare1}, and 
 Fig.~\ref{fig:nbn_lon_compare2}, we can find several features:
\begin{itemize}
    \item Ruggedness
    
    The TSP instance exhibits ruggedness from global to local regions with numerous straight hanging points. While in Fig.~\ref{fig:nbn_tsp_substructure}, Fig.~\ref{fig:nbn_lon_compare1}, and 
 Fig.~\ref{fig:nbn_lon_compare2},  we can see that whether it's NBN with LON data or NBN with total data, the fitness landscape has been considerably smoothed out. This indicates that local search operators can smoothen the structure of the TSP fitness landscape, which also validates the importance of local search operators for TSP.

  \item Modality
  
From the results in Fig.~\ref{fig:nbn_tsp_substructure}, 
 we can see that \text{rue500-1} exhibits a single BoA globally,  with multiple BoAs emerging in the local structure when $K$ is smaller than 50. In the local structure of $K = 12$, there are two connected BoAs. 

\end{itemize}

\begin{figure*}[!t]
\centering
  \begin{tabular}{|cccc|}
  \hline
  \multicolumn{4}{|c|}{u574}                                                                       \\ \hline
  \multicolumn{2}{|c|}{LKH's success, $\boldsymbol{T}_0$} & \multicolumn{2}{c|}{EAX's success, $\boldsymbol{T}_0$} \\
   \includegraphics[width=0.2\textwidth]{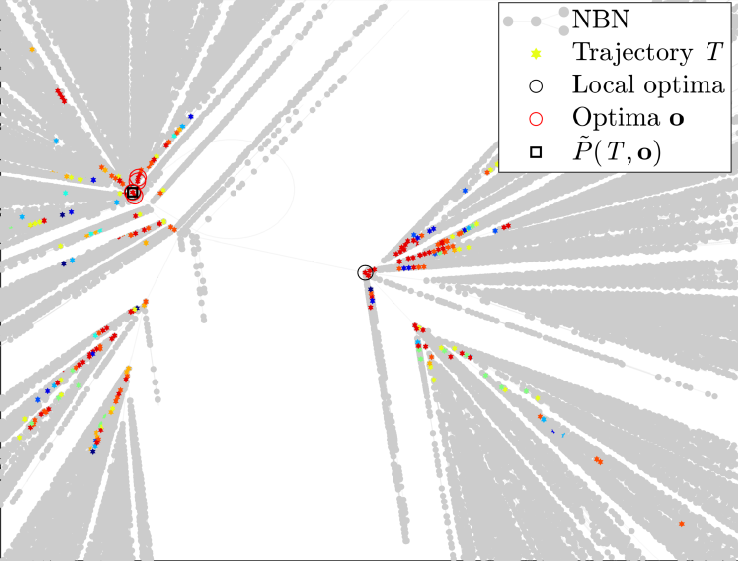}      
   & \multicolumn{1}{c|}{
    \includegraphics[width=0.2\textwidth]{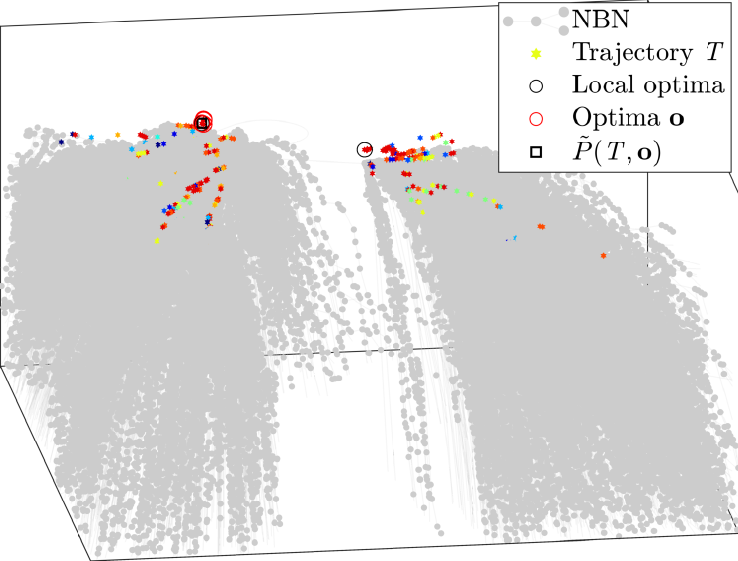} }
   & \includegraphics[width=0.2\textwidth]{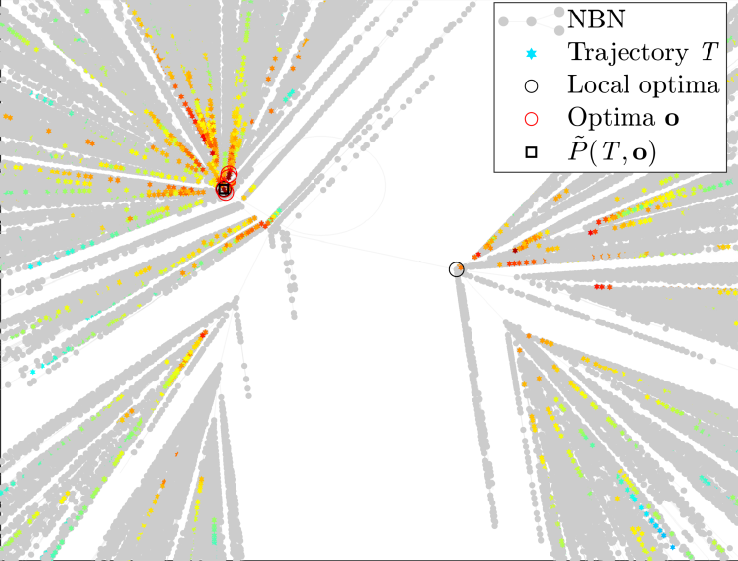}        & \includegraphics[width=0.2\textwidth]{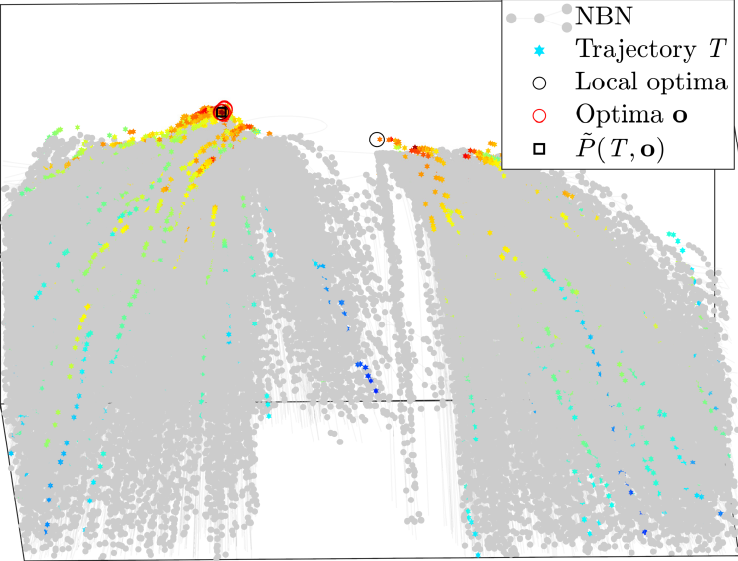}                       \\
  top view    & \multicolumn{1}{c|}{side view}    & top view              & side view              \\ \hline
  \multicolumn{4}{|c|}{rue500-1}                                                                   \\ \hline
  \multicolumn{2}{|c|}{LKH's success, $\boldsymbol{T}_0$} & \multicolumn{2}{c|}{EAX's success, $\boldsymbol{T}_{12}$} \\
  \includegraphics[width=0.2\textwidth]{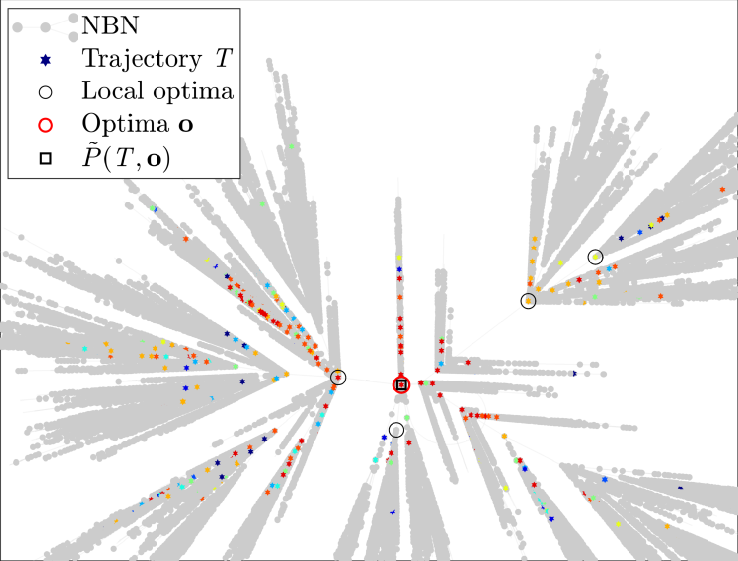}      
   & \multicolumn{1}{c|}{
    \includegraphics[width=0.2\textwidth]{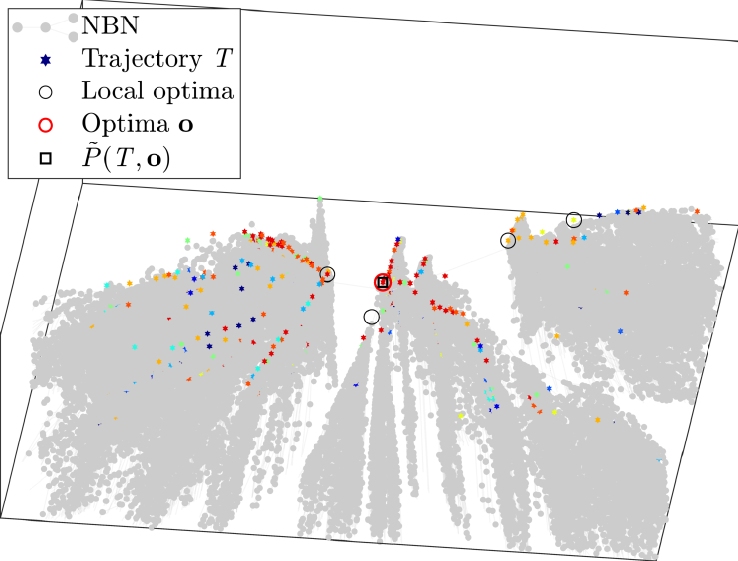} }
   & \includegraphics[width=0.2\textwidth]{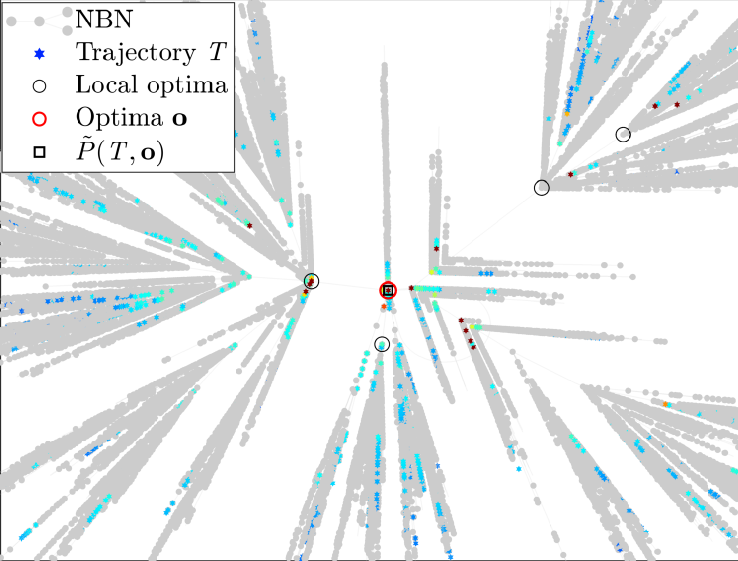}        & \includegraphics[width=0.2\textwidth]{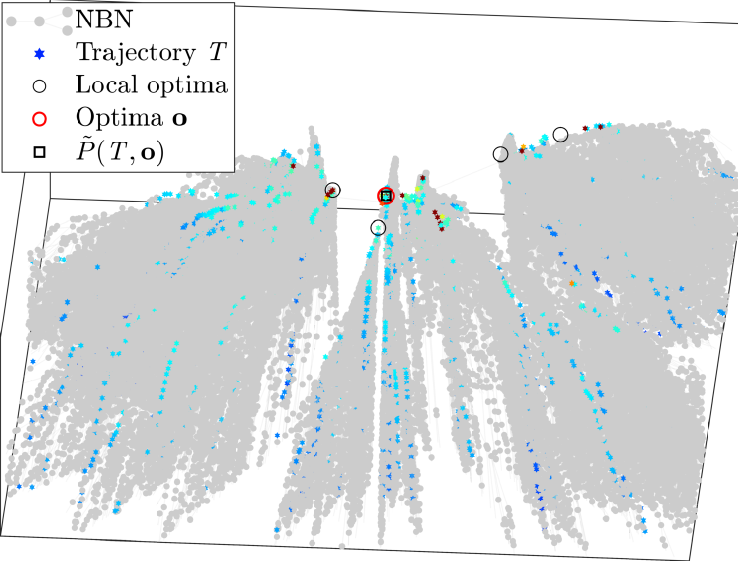}                       \\
  top view    & \multicolumn{1}{c|}{side view}    & top view              & side view              \\ \hline

  \multicolumn{2}{|c|}{EAX's failure, $\boldsymbol{T}_9$, $d(\boldsymbol{T}_9, \mathbf{o}) = 3$} & \multicolumn{2}{c|}{EAX's failure, $\boldsymbol{T}_{15}$, $d(\boldsymbol{T}_{15}, \mathbf{o}) = 17$} \\
  \includegraphics[width=0.2\textwidth]{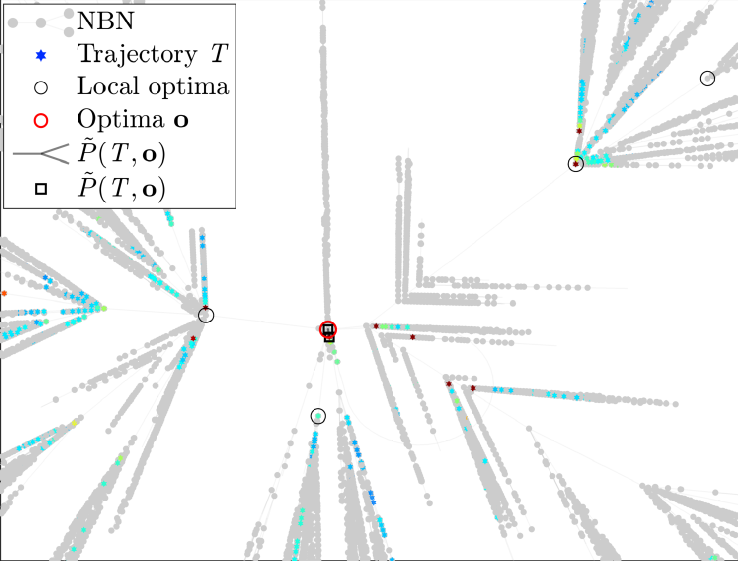}      
   & \multicolumn{1}{c|}{
    \includegraphics[width=0.2\textwidth]{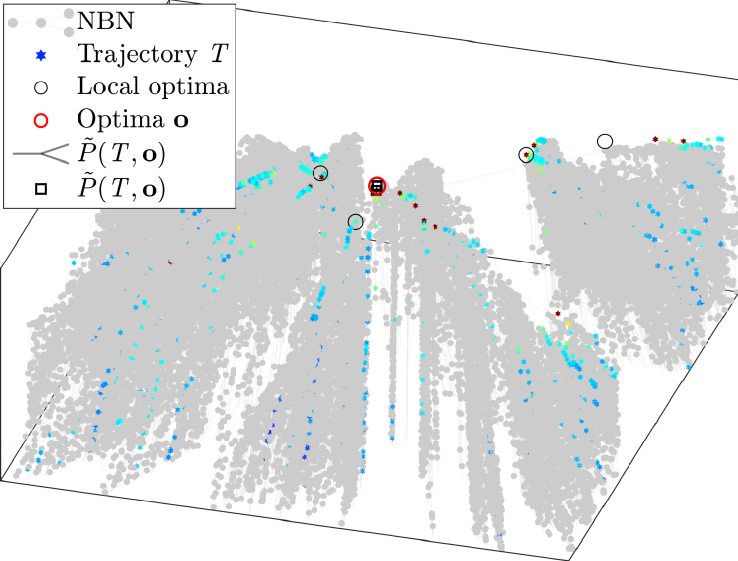} }
   & \includegraphics[width=0.2\textwidth]{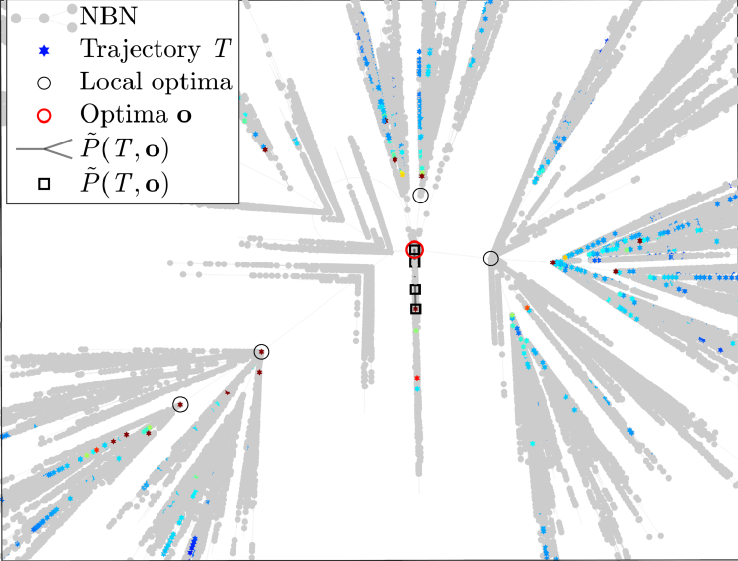}        & \includegraphics[width=0.2\textwidth]{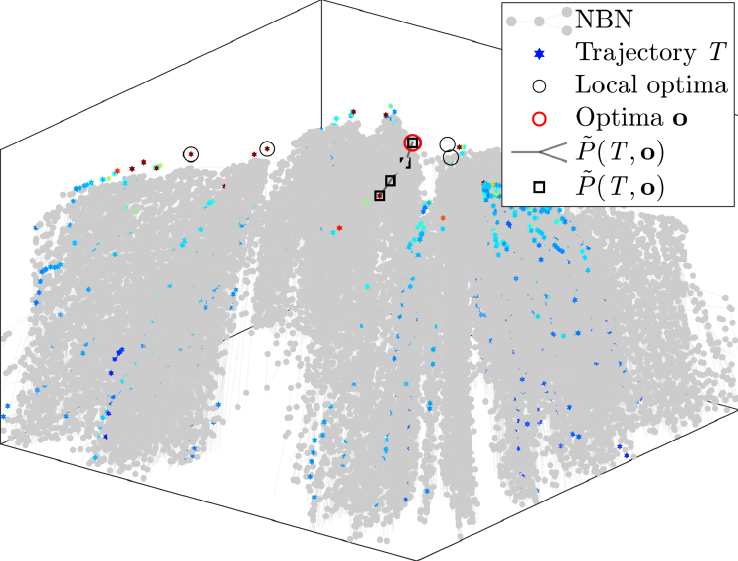}                       \\
  top view    & \multicolumn{1}{c|}{side view}    & top view              & side view              \\ \hline

  \multicolumn{4}{|c|}{rue500-2}                                                                   \\ \hline
  \multicolumn{2}{|c|}{LKH's success, $\boldsymbol{T}_3$} & \multicolumn{2}{c|}{EAX's success, $\boldsymbol{T}_{12}$} \\
  \includegraphics[width=0.2\textwidth]{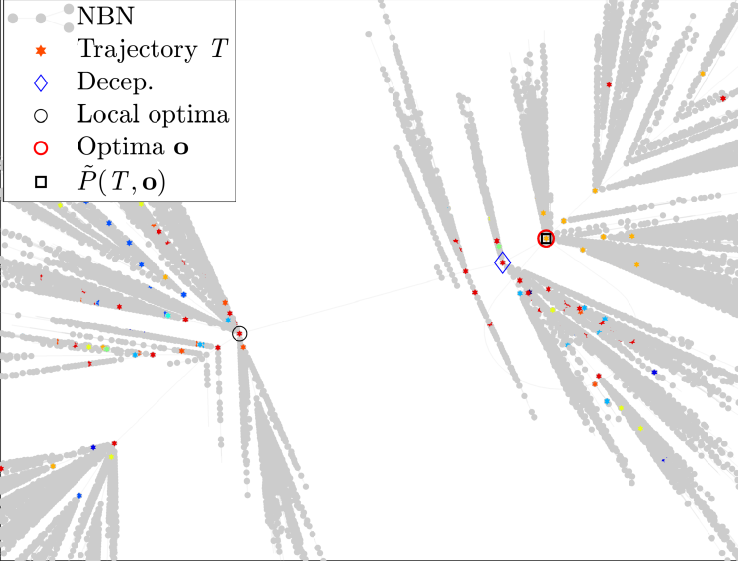}      
   & \multicolumn{1}{c|}{
    \includegraphics[width=0.2\textwidth]{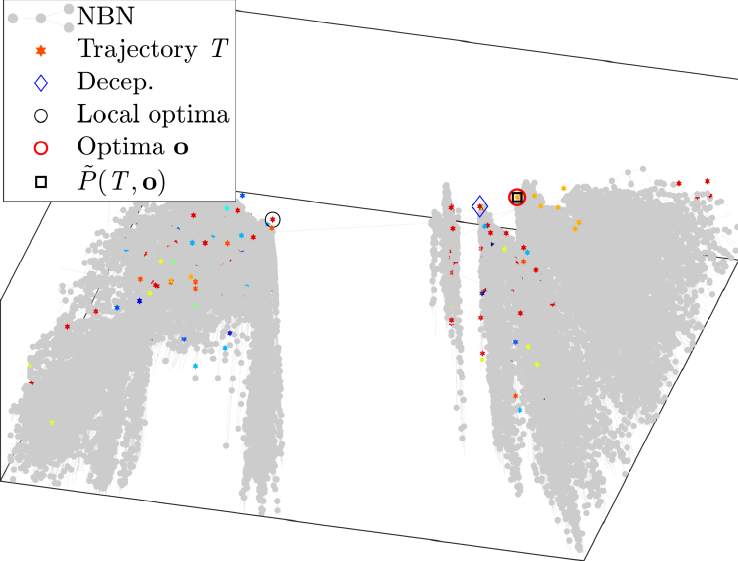} }
   & \includegraphics[width=0.2\textwidth]{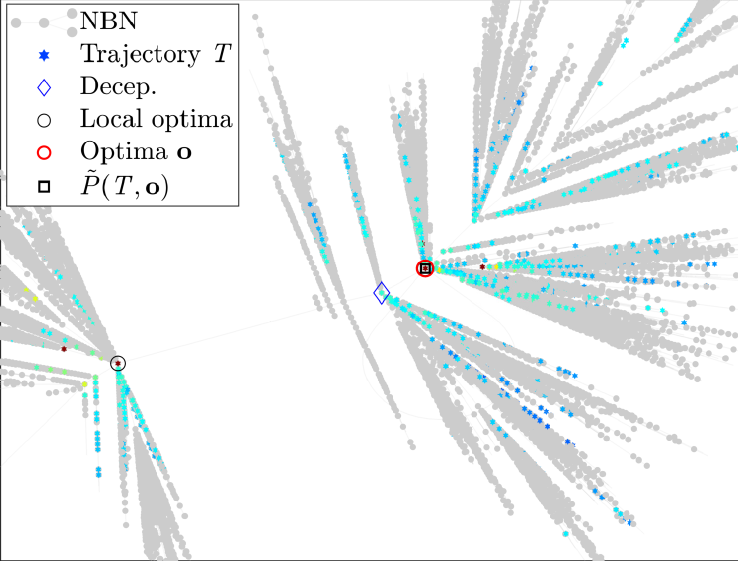}        & \includegraphics[width=0.2\textwidth]{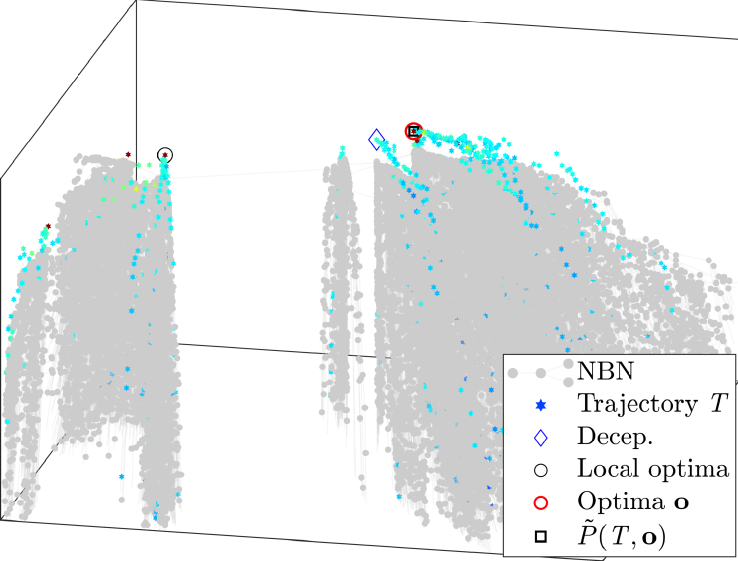}                              \\
  top view    & \multicolumn{1}{c|}{side view}    & top view              & side view              \\ \hline
   
  \multicolumn{2}{|c|}{LKH's failure, $\boldsymbol{T}_0$, $d(\boldsymbol{T}_0, \mathbf{o}) = 15$} & \multicolumn{2}{c|}{LKH's failure, $\boldsymbol{T}_{5}$, $d(\boldsymbol{T}_{5}, \mathbf{o}) = 7$} \\
  \includegraphics[width=0.2\textwidth]{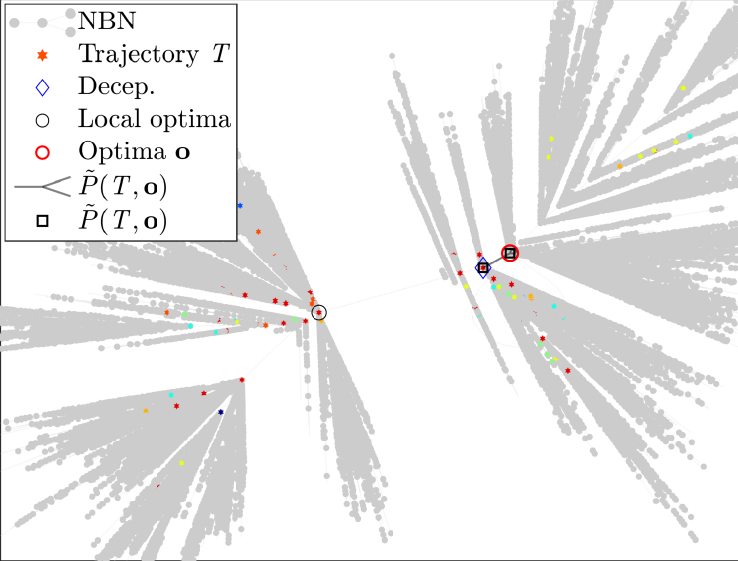}      
   & \multicolumn{1}{c|}{
    \includegraphics[width=0.2\textwidth]{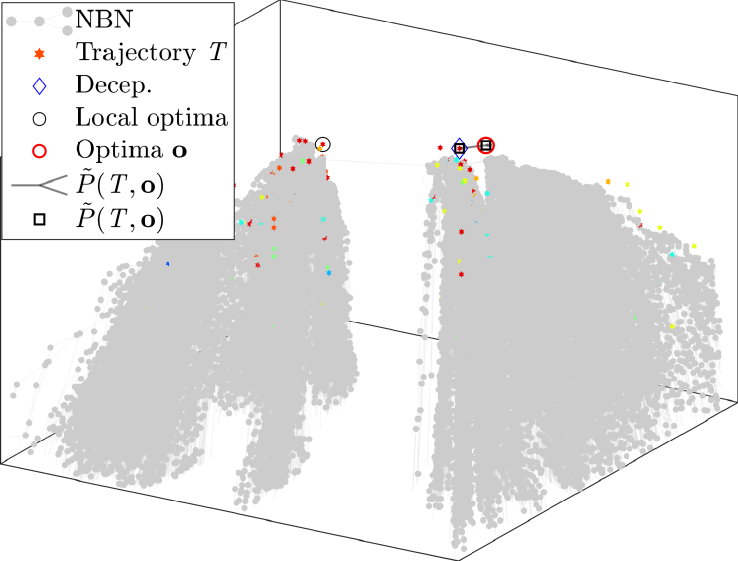} }
   & \includegraphics[width=0.2\textwidth]{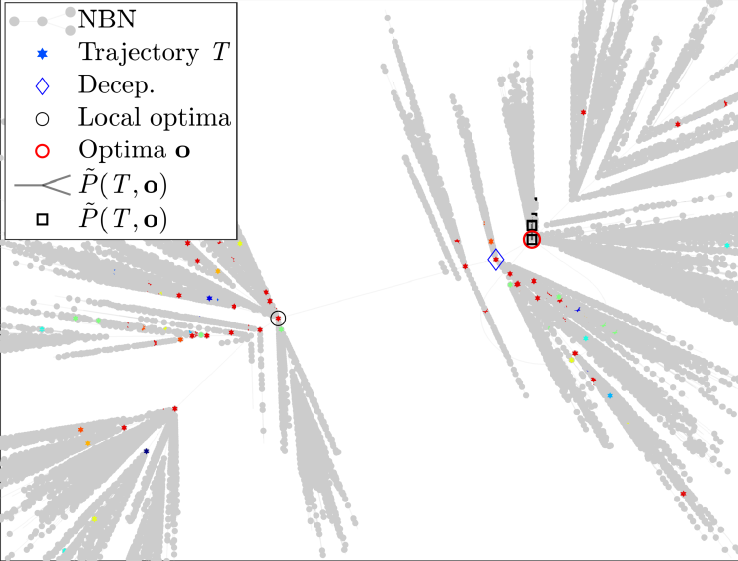}        & \includegraphics[width=0.2\textwidth]{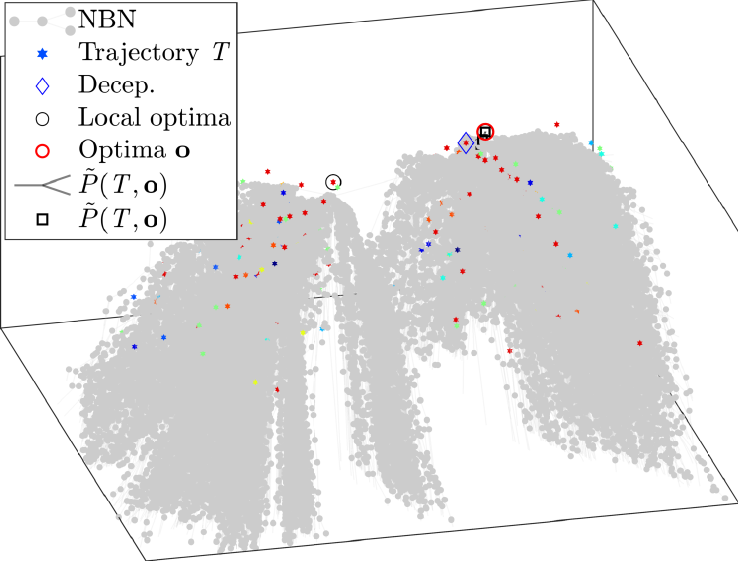}                         \\
  top view    & \multicolumn{1}{c|}{side view}    & top view              & side view              \\ \hline
  \end{tabular}
    \caption{NBN visualization with total data, where the gray network is NBN, colored stars are the solutions of an algorithm's trajectory $\boldsymbol T$.
    $\boldsymbol T_i$ indicates the $i^{th}$ trajectory of an algorithm.
    Black circles indicate the local optima and red circles are the global optima $\mathbf o$.  Diamond markers indicate deceptive solutions.
    Black rectangles are the solutions along the shortest evolutionary path from the trajectory $\boldsymbol T$ to the global optima $\tilde{P}(\boldsymbol T, \mathbf o) $.
    $d(\boldsymbol{T}, \mathbf{o})$ is the distance of the evolutionary path. For EAX's trajectories, the color represents the generated iteration of the solutions.
    For LKH's trajectories, the color represents the number of runs that the solutions are generated.
    }
    \label{fig:nbn_alg_trait}
  
  \end{figure*}

\subsubsection{Comparison between LON and NBN}

For the same dataset, the structures of LON and NBN exhibit similarities as shown in Fig\:\ref{fig:nbn_lon_compare1} and  Fig\:\ref{fig:nbn_lon_compare2}. Due to the high time complexity of the force-directed algorithm used in LON ($N^2$, where $N$ is the number of solutions), it can only visualize the best 0.01\% of solutions. 
The similarities between LON and NBN suggest that although NBN only preserves the nearest-better relationship within the network, it still retains the features of BoAs.

 LON relies on local search operators to explore relationships between solutions, which smoothens the fitness landscape, 
 whereas NBN can visualize solutions from any source and retains important features such as ruggedness as illustrated in Fig.~\ref{fig:nbn_tsp_substructure}.  
 The experiments below also show that NBN provides more information for a more profound analysis of the problem features and algorithm behaviors.
 
\subsubsection{Algorithm behavior analysis} 

To compare the capabilities of LON, LDEE, and NBN in analyzing algorithm's behavior, we evaluate TSP instances using the data generated by each tool individually.

 \begin{itemize}
  \item Analysis based on LON

LON can display the connections between local optima and further illustrate the BoAs. As illustrated in Fig.~\ref{fig:nbn_lon_compare1} and  Fig.~\ref{fig:nbn_lon_compare2}, all three TSP instances have two BoAs, and in rue500-1, most regions of the BoAs are connected. It seems that rue500-1 is easier than the other two instances.  But is that correct? From the results in \text{TABLE\:\ref{tb:eax_lkh_success}}, LKH has a low success rate on rue500-2. LON does not provide a clear answer as to whether the algorithm consistently gets stuck in the red local optima, as shown in \text{Fig.~\ref{fig:nbn_lon_compare2}}.

  On the other hand, both u574 and rue500-2 have two separate BoAs, yet EAX consistently finds the global optimum on the two instances.  However, in rue500-1, which appears simpler with two connected BoAs, the algorithm's success rate is quite low. Based on LON visualization, it is difficult to identify the specific factors causing the poor performance of EAX in this instance.

  \item Analysis based on LDEE

  LDEE maps all solutions onto a two-dimensional plane by minimizing the total distance between each pair of solutions. 
  From the visualization, we can see that much information is lost, making it difficult to infer details about the landscape features. We can only observe the relative positions of the solutions and their fitness values.
 
LDEE focuses on minimizing the overall relative distances between all pairs of solutions, which may cause the loss of some critical information. 
The information about the distance between optima is of great importance to the researcher. 
As shown in Fig.~\ref{fig:nbn_lon_compare1}, in NBN of u574, four optima are very close to each other. In LON of u574, the four optima are also mutually reachable.
In the LDEE visualization with total data, the positions of the four optima are close. However, in the LDEE visualization with LON and EAX data, we can see two sets of distant optima (white rectangles), which can be misleading, since one may believe that there are two groups of optimal solutions far apart in u574. The difference between LDEE with total data and with LON, EAX data also indicates that LDEE's mapping involves a significant degree of randomness.

Furthermore, it is hard to draw any effective conclusions about algorithm behavior from the LDEE visualization. 
In the case of EAX's failure on rue500-1, we can see that EAX finds some solutions very close to the global optimum. 
But despite this proximity, why does EAX fail to find the global optimum? LDEE does not provide answers to this question.

Similarly, in the case of LKH's failure in rue500-2, LDEE shows that the solutions generated by LKH are close to the global optimum as shown in Fig.~\ref{fig:nbn_lon_compare2}. It seems that LKH with the nearest solution can converge to the global optimum using any local search operator. 
But is this the case? The following NBN-based analysis shows that in this trajectory $\boldsymbol T_0$, LKH gets stuck in a deceptive funnel.

\item Analysis based on NBN

To better analyze the behaviors of the two algorithms, 
different trajectories are shown in \text{Fig.\:\ref{fig:nbn_alg_trait}}.
The statistical information of $d (\tilde{P}(\boldsymbol T, \mathbf o)) $ is shown in \text{TABLE\:\ref{tb:d_P_data}}, where ``Failed TSP instance''  indicates the instances where they have a low success rate. ``min'', ``max'', ``mean'', and ``SD''  represent the minimum value, maximum value, mean value, and standard deviation of  $d (\tilde{P}(\boldsymbol T, \mathbf o)) $, respectively.   `` Fails (Deceptive/Total)'' indicates the number of failures when the algorithm gets stuck in deceptive solutions versus the total number of failures.

  \begin{table*}[!htbp]\centering
    \caption{Statistical Information of $d (\tilde{P}(\boldsymbol T, \mathbf o))  $ over the 30 independent trajectories for the two algorithms on their failed TSP instances. } 
    \label{tb:d_P_data}
    \begin{tabular}{|c|c|c|c|c|c|}
    \hline
    Algorithm & Failed TSP instance      & min & max & Avg ± SD         & Fails (Deceptive/Total) \\ \hline
    EAX       & rue500-1 & 3   & 17  & $9.25 \pm  3.94267$    & -/28                       \\ \hline
    LKH       & rue500-2 & 7   & 15  & $14.6923 \pm  1.53846$ & 25/26                   \\ \hline
    \end{tabular}
    \end{table*}

\begin{table*}[t]
  \centering
  \caption{ Statistic data of the funnels around optima, $K = 16$  }
  \label{tb:funnel_data}
  \begin{tabular}{|c|c|c|c|c|cc|cc|cc|}
  \hline
  \multirow{2}{*}{}         & \multirow{2}{*}{Id} & \multirow{2}{*}{Decep.} & \multirow{2}{*}{$\|\mathbf n, \mathbf o\|$} & \multirow{2}{*}{NBD} & \multicolumn{2}{c|}{$N=10,000$}               & \multicolumn{2}{c|}{$N=100,000$}              & \multicolumn{2}{c|}{$N=1,000,000$}             \\ \cline{6-11} 
                            &                         &                            &                    &                      & \multicolumn{1}{c|}{$\Delta  \overline{f}  $}                &  $ d(\boldsymbol S(\mathbf n, K), \mathbf{o})$   & \multicolumn{1}{c|}{$\Delta  \overline{f} $}                   &  $ d(\boldsymbol S(\mathbf n, K), \mathbf{o})$  & \multicolumn{1}{c|}{$\Delta  \overline{f} $}                   &  $ d(\boldsymbol S(\mathbf n, K), \mathbf{o})$  \\ \hline
  \multirow{3}{*}{u574}     & 1                       &                            & 15                 & 11                   & \multicolumn{1}{c|}{6.86E-03}           & 37 & \multicolumn{1}{c|}{6.63E-03}           & 34 & \multicolumn{1}{c|}{7.26E-03}           & 34 \\ \cline{2-11} 
                            & 2                       &                            & 17                 & 12                   & \multicolumn{1}{c|}{9.09E-03}           & 40 & \multicolumn{1}{c|}{5.49E-03}           & 41 & \multicolumn{1}{c|}{5.19E-03}           & 45 \\ \cline{2-11} 
                            & 3                       &                            & 17                 & 13                   & \multicolumn{1}{c|}{\textbf{-1.47E-02}} & 38 & \multicolumn{1}{c|}{\textbf{-1.04E-02}} & 41 & \multicolumn{1}{c|}{\textbf{-1.08E-02}} & 32 \\ \hline
  \multirow{2}{*}{rue500-1} & 1                       &                            & 12                 & 12                   & \multicolumn{1}{c|}{2.57E-03}           & 32 & \multicolumn{1}{c|}{6.85E-04}           & 37 & \multicolumn{1}{c|}{1.88E-04}           & 34 \\ \cline{2-11} 
                            & 2                       &                            & 17                 & 10                   & \multicolumn{1}{c|}{2.56E-03}           & 41 & \multicolumn{1}{c|}{7.59E-04}           & 38 & \multicolumn{1}{c|}{1.28E-03}           & 37 \\ \hline
  \multirow{5}{*}{rue500-2} & 1                       &                            & 17                 & 14                   & \multicolumn{1}{c|}{1.73E-02}           & 44 & \multicolumn{1}{c|}{1.69E-02}           & 39 & \multicolumn{1}{c|}{1.63E-02}           & 36 \\ \cline{2-11} 
                            & 2                       &                            & 11                 & 11                   & \multicolumn{1}{c|}{8.14E-04}           & 36 & \multicolumn{1}{c|}{2.98E-04}           & 34 & \multicolumn{1}{c|}{7.06E-04}           & 33 \\ \cline{2-11} 
                            & 3                       &                            & 13                 & 13                   & \multicolumn{1}{c|}{\textbf{-1.83E-03}} & 41 & \multicolumn{1}{c|}{5.81E-04}           & 36 & \multicolumn{1}{c|}{7.82E-04}           & 32 \\ \cline{2-11} 
                            & 4                       & $\surd$                      & 15                 & 15                   & \multicolumn{1}{c|}{\textbf{-1.43E-03}} & 38 & \multicolumn{1}{c|}{\textbf{-4.43E-04}} & 35 & \multicolumn{1}{c|}{\textbf{-6.69E-04}} & 30 \\ \cline{2-11} 
                            & 5                       &                            & 16                 & 12                   & \multicolumn{1}{c|}{3.10E-04}           & 40 & \multicolumn{1}{c|}{4.18E-04}           & 36 & \multicolumn{1}{c|}{3.71E-04}           & 31 \\ \hline
  \end{tabular}
  \end{table*}

\begin{itemize}
    \item EAX's  behaviors

From \text{TABLE\:\ref{tb:eax_lkh_success}}, we see that EAX  fails only on rue500-1. In \text{Fig.\:\ref{fig:nbn_alg_trait}}, we observe that solutions exist within the BoAs of the global optima in both success and failure cases. \textcolor{version_3}{
 Even in trajectory $\boldsymbol T_{15}$  when the distance to the optima is relatively large, $d (\tilde{P}(\boldsymbol T_{15}, \mathbf o)) = 17$, there still exist several solutions in BoA of the global optima. 
 EAX does not suffer from a lack of diversity in detecting the BoA of the optimum. On the contrary, it successfully locates the BoA of the global optima. }

 \textcolor{version_3}{
Moreover, \text{TABLE\:\ref{tb:d_P_data}} shows that $d (\tilde{P}(\boldsymbol T, \mathbf{o}))$ varies significantly across different trajectories, suggesting that the algorithm converges to different locations in these trajectories,
which also validates the ability of EAX to maintain diversity.
}

\textcolor{version_3}{
Although EAX can detect the BoA of global optima in all these trajectories, it still has a low success rate on \text{rue500-1} as shown in TABLE\:\ref{tb:eax_lkh_success}. Then, we further analyze EAX's behavior on \text{rue500-1}.} In the $9^{th}$ trajectory $\boldsymbol T_9$, the distance between $\boldsymbol T_9$ and the global optima $\mathbf{o}$ is only 3. \textcolor{version_3}{
It seems that the local search operators applied to the nearest solution could easily find the global optimum. However, EAX relies on edge assembly crossover to improve the solutions.  When multiple BoAs exist, EAX retains individuals within multiple basins simultaneously, reducing inter-basin interaction efficiency and leading to algorithm's stagnation.
The result of the optima screened according to Eq.~\eqref{eq:num_opt}  with $\theta = 0.99$ and $\vartheta = 30$ also supports this conclusion, where 
\text{rue500-1} has 5 optima, while the other two instances have only 2 optima. All the optima are marked as circles in \text{Fig.\:\ref{fig:nbn_alg_trait}}.
}

    \item LKH's  behaviors

As shown in \text{TABLE\:\ref{tb:eax_lkh_success}}, LKH struggles with \text{rue500-2}. Compared to the EAX's behaviors, we know that modality is not the challenge that LKH encounters. 

In \text{Fig.\:\ref{fig:nbn_alg_trait}}, we can see that in \text{rue500-2} LKH tends to converge to the local optima and the blue diamond-shaped solution. Even in the LKH's successful trajectory, $\boldsymbol T_3$, there are many solutions around the blue diamond-shaped solution. Only one set of solutions (orange stars) generated in the same run are situated around the global optima. Besides, EAX also has many solutions around the blue diamond-shaped solution. It seems that EAX treats it like a local optimum.

The data in \text{TABLE\:\ref{tb:d_P_data}} also corroborates this phenomenon. 
Among the 26 failures, LKH is stuck in the blue diamond-shaped solution 25 times. Then, is the solution deceptive? To verify this hypothesis, we need to answer two questions: (1) Why is that the other two instances do not have deceptive solutions? (2) Why is the blue diamond-shaped solution the only deceptive solution in \text{rue500-2}?

We know that a deceptive solution is a solution that is close to the global optimum with better BoAs, so algorithms are attracted by the deceptive solution and thus ignore the global optima.  Based on this, we filter out all the potential deceptive solutions in all three instances as shown in \text{TABLE\:\ref{tb:funnel_data}}. 
 The largest local search operator used by LKH is the 5-opt, which indicates that LKH can find the best solutions in a local area with a radius $K \leq 10$.
 \textcolor{version_3}{
 For a solution with NBD smaller than  10, LKH can find its nearest better solution using the 5-opt local search operator.
 }
 Thus, any possible deceptive solution should have an NBD larger than 10 so that LKH can converge to it instead of the global optimum. 
 Additionally, the deceptive solution should be closer to the global optimum, so that it can shadow the global optimum for LKH. 
 Thus, we filter out all the possible potential solutions based on the following metric:
 \begin{equation}
    \label{eq:decept_metric}
    d_{\mathrm{NBD}} (\mathbf{n}) \geq 10 \, \land \,      \|\mathbf n, \mathbf o\|  \leq 17
    \end{equation}

Next, we analyze the local structure around the potential deceptive solution and the global optimum.
We performed local sampling around both the deceptive solution and the global optimum, resulting in two solution sets, $\boldsymbol S (\mathbf n, K) $ and  $\boldsymbol S( \mathbf o, K)$ with a sampling radius of $K = 17$. Then, we analyze the difference of the average fitness of the two solution sets, denoted as 
$\Delta  \overline{f} $ in  \text{TABLE\:\ref{tb:funnel_data}}, calculated using the following formula: 
   \begin{equation}
    \label{eq:delta_f}
    \Delta  \overline{f} =  \frac{\sum_{\mathbf x_i \in \boldsymbol S (\mathbf o, K)  } f(\mathbf x_i)  }{N}  -\frac{\sum_{\mathbf x_i \in \boldsymbol S (\mathbf n, K)  } f(\mathbf x_i)  }{N} 
   \end{equation}

  \begin{figure*}[!t]\centering
    \subfloat[u574, Id = 3]{\label{fig:funnel_u574_ID3}
    \begin{tabular}[b]{cc}
    \includegraphics[width=.225\linewidth]{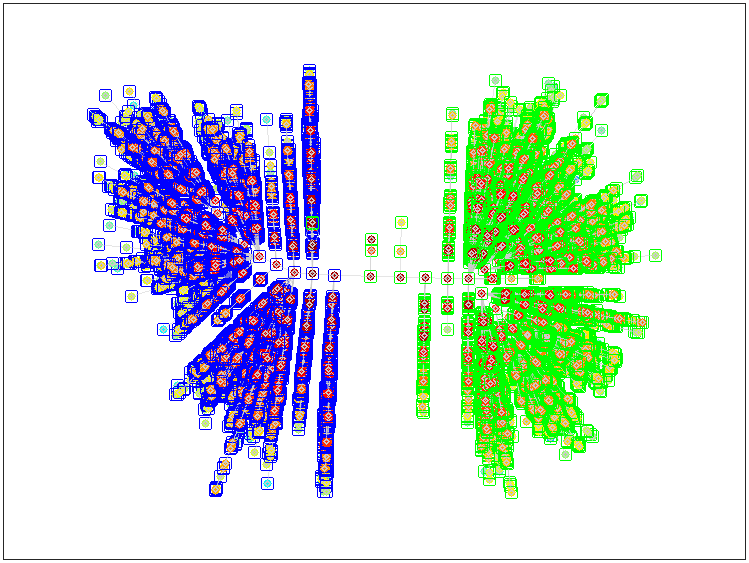} &
    \includegraphics[width=.225\linewidth]{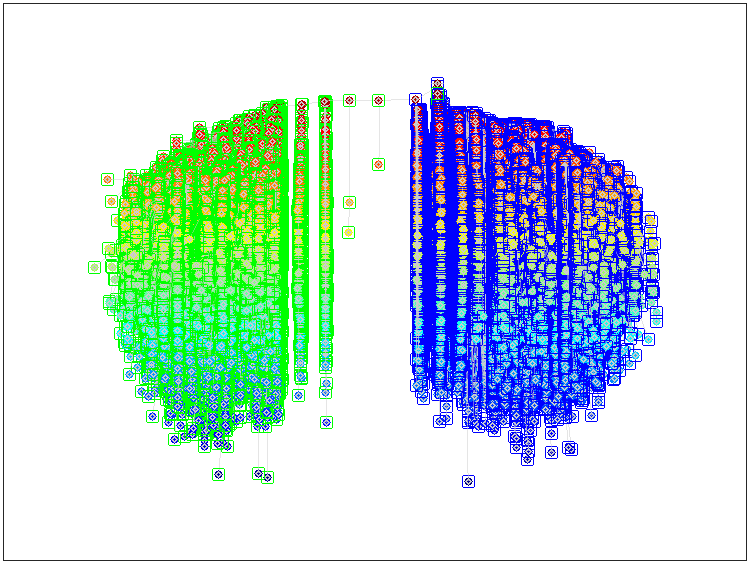}\\
     top view  & side view
    \end{tabular}%
    }
    \subfloat[rue500-2, Id = 4]{\label{fig:funnel_rue500_2_ID4}
    \begin{tabular}[b]{cc}
    \includegraphics[width=.225\linewidth]{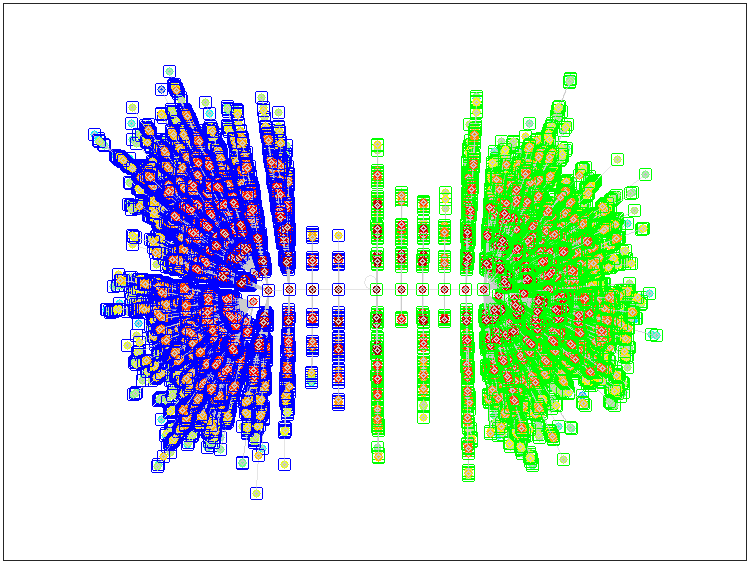} &
    \includegraphics[width=.225\linewidth]{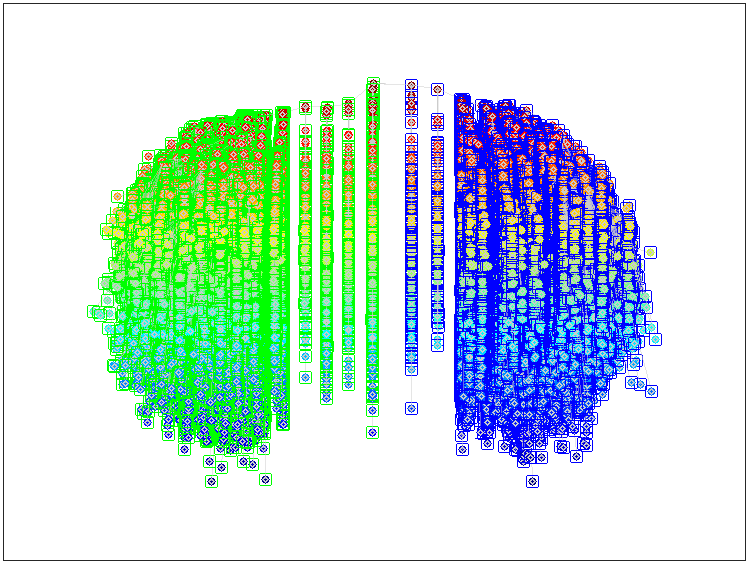}\\
    top view  & side view
    \end{tabular}%
    }
    \caption{NBN of the combined data of $N=10^6$ solutions from local sampling with a radius of $K = 17$ around the possible deceptive solution and global optima, where the blue rectangles represent the solutions around the possible deceptive solution and the green rectangles represent the solutions around the optima. }\label{fig:NBN_decpt_two_funnel}
    \end{figure*}

   From \text{TABLE\:\ref{tb:funnel_data}}, we observe that only two funnels have better local areas than the global optimum: Funnel 3 of u574 and Funnel 4 of \text{rue500-2}. Interestingly,  Funnel 4 of \text{rue500-2} is the deceptive solution that we predicted, i.e., the blue diamond-shaped solution in \text{Fig.\:\ref{fig:nbn_alg_trait}}. 
   This indicates that both funnels are potentially deceptive solutions.

  We need to further analyze whether the BoAs of the two funnels are close to the BoA of the optima so that the solutions in the BoA of the global optimum are easily attracted by the funnel.
   By analyzing the NBN of the combined  data of the two solution sets, as shown in \text{TABLE\:\ref{tb:funnel_data}}, 
   we found out that the sampled solution set of funnel 4 of \text{rue500-2} is closer to the global optimum compared to funnel 3 of \text{u574}, with a smaller $ d(\boldsymbol S(\mathbf n, K), \mathbf{o})$. 
This distance is even shorter when compared to some other funnels, particularly in a larger sampled solution set ($N = 1e6$). Furthermore, 
   \text{Fig.~\ref{fig:NBN_decpt_two_funnel}} also shows that many solutions connect the two funnels in NBN of \text{rue500-2, Id = 4} than in \text{u574, Id = 3}.
This suggests that the solutions around the global optimum in \text{rue500-2} are more easily attracted to funnel 4.

Note that the NBN created using the two local sampling datasets is biased, with few solutions located at the center of the funnel and the global optima. Therefore, the distance 
 $ d(\boldsymbol S(\mathbf n, K), \mathbf{o})$  may not be accurate, and the true value is likely smaller.  However, for different funnels, the distribution of the sampled solutions remains consistent, making the distances 
 $ d(\boldsymbol S(\mathbf n, K), \mathbf{o})$ of different funnels comparable.
    
\end{itemize}
\end{itemize}

\subsubsection{Conclusions from the Analysis}

Based on the NBN-assisted analysis, three primary challenges are identified for TSP: ruggedness,  modality, and deception.  Most TSP local search operators can overcome ruggedness challenges, as NBN structures generated from optimization data are remarkably smoother compared to randomly sampled NBN structures. 

EAX struggles with the modality challenge. EAX can efficiently maintain diversity, but when there are many optima in the problem, solutions are distributed in different BoAs. Few solutions are located in the BoAs of the global optima and, therefore, it is hard to converge to the global optima. While LKH does not suffer from the modality challenge. It relies on its local search operators and can converge to different optima with a random restart in each run. LKH struggles with the deception challenge. When there is a deceptive solution near the global optima, the local-search-based LKH tends to converge to the deceptive solution. While EAX just treats it as a local optimum. 

 \section{Conclusions and Future Work}\label{sec:conclusion}

 In this paper, we offered a straightforward proof indicating that NBN fundamentally represents the maximum probability transition network. We also presented an efficient calculation method for NBN with logarithmic linear time complexity for assignment problems. Furthermore, we conducted an in-depth analysis in OneMax problems and TSP. For the first time, we found that the fitness landscape of OneMax exhibits neutrality, ruggedness, and modality features.
 We also uncovered some limitations of the state-of-the-art TSP algorithms: 
 LKH, which relies on its local search operators, fails when there are deceptive solutions near the global optima. While, EAX, based on a single population, efficiently maintains diversity. However, when multiple attraction basins exist, it retains individuals within multiple basins simultaneously, reducing inter-basin interaction efficiency and leading to algorithm's stagnation as well.

 We believe that since NBN can reveal the underlying challenges of the problems, it can also solve them.  To tackle the limitations of current TSP algorithms in dealing with modality and deception challenges, we aim to develop NBN-based algorithms capable of adaptively learning these landscape features.

\bibliographystyle{IEEEtran}
\bibliography{ref}

%



%







\end{document}